%% file: main.tex
\documentclass[british]{article}
\usepackage[utf8]{inputenc}
\usepackage{amsfonts, amsmath, amssymb, amsthm, bbm, color}
\usepackage{mathrsfs}           
\usepackage{xcolor, colortbl}
\usepackage{u8chars}
\usepackage{relsize}
\usepackage{xspace}
\usepackage{comment}            
\usepackage[inline,shortlabels]{enumitem}
\usepackage[group-separator={,},group-four-digits]{siunitx}
\usepackage{url}
\usepackage[numbers]{natbib}
\usepackage[textsize=small]{todonotes}
\usepackage{tikz}
\usepackage{graphicx}
\usepackage[%
colorlinks,
linkcolor={red!50!black},
citecolor={blue!50!black},
urlcolor={blue!80!black},
]{hyperref}

\title{Abstraction and Symbolic Execution of Deep Neural Networks with Bayesian Approximation of Hidden Features}
\author{Nicolas Berthier$^1$, Amany Alshareef$^1$, James Sharp$^2$, \\Sven Schewe$^1$, Xiaowei Huang$^1$}
\date{$^1$ University of Liverpool, UK \\
$^2$ Defence Science and Technology Laboratory (Dstl), UK}

\input{macro.tex}
\input{mylists.tex}

\input{tikzconf.tex}

\newtheorem{lemma}{Lemma}

\theoremstyle{definition}
\newtheorem{defn}{Definition}[section]
\newtheorem{example}{Example}[section]

\excludecomment{leaveout}

\begin{document}

\maketitle

\begin{abstract}
  Intensive research has been conducted on the verification and validation of deep neural networks (DNNs), aiming to understand if, and how, DNNs can be applied to safety critical applications.
  However, existing verification and validation techniques are limited by their scalability, over both the size of the DNN and the size of the dataset.
  In this paper, we propose a novel abstraction method which abstracts a DNN and a dataset into a Bayesian network (BN).
  We make use of dimensionality reduction techniques to identify hidden features that have been learned by hidden layers of the DNN, and associate each hidden feature with a node of the BN.
  On this BN, we can conduct probabilistic inference to understand the behaviours of the DNN processing data.
  More importantly, we can derive a runtime monitoring approach to detect in operational time rare inputs and covariate shift of the input data.
  We can also adapt existing structural coverage-guided testing techniques (\ie based on low-level elements of the DNN such as neurons), in order to generate test cases that better exercise hidden features.
  We implement and evaluate the BN abstraction technique using our DeepConcolic tool available at \url{https://github.com/TrustAI/DeepConcolic}.
\end{abstract}

\section{Introduction}

Neural networks generally work with high precision, but recent work has shown that they are subject to weaknesses such as adversarial attacks \cite{szegedy2014intriguing}, data poisoning attacks \cite{10.5555/3042573.3042761}, Trojan attacks \cite{LMALZWZ2018}, model inversion attacks \cite{FJR2015}, etc.
Given their importance and such weaknesses, the analysis of DNNs has become a popular research direction, with research on formal verification, coverage-guided testing, etc. See \cite{HUANG2020100270} for a recent survey.
The large size of DNNs, containing tens of thousands of neurons that interact with each other in intricate ways, leads to the scalability problem of these analysis methods, particularly for white-box analysis techniques, whose computational complexity is usually measured over either the number of neurons or the number of parameters.
Moreover, most analysis methods -- either verification or testing -- work with local inputs individually, which easily leads to the other scalability problem when they need to work with a large dataset.
The above scalability problems, due to the size of DNNs and the size of datasets, imply the need for an abstraction technique that approximates a DNN and a dataset into a simpler model -- while preserving critical properties -- to improve the scalability of the analysis methods.

\paragraph{\bf Contribution.}
This paper proposes a novel abstraction technique for DNNs through Bayesian approximation: we abstract the behaviour of a DNN on a dataset into a Bayesian network (BN), which is a probabilistic graphical model based on high-level features instead of low-level neurons.

The first step of this abstraction is naturally to identify the relevant \emph{hidden features} for each hidden layer.
Standard feature extraction techniques are available, and while we obviously have to select some for experimental results and explore a few different ways of conducting feature extraction, we predominantly treat feature extraction as black-box techniques.

The features not only provide the structure of the BN, they also provide, for each input from a given dataset\footnote{For most methods, the dataset that has been used to learning a DNN is also used to identify the features. But we also use fixed features, e.g.\ when a DNN is used over a new dataset, and we are also keen to keep this part of the analysis independent from how the features are selected.}, an observation of occurrences of features in neighbouring layers.

This first allows us to conduct probabilistic reasoning to understand how the appearance of some input features may statistically affect the appearance of hidden features or output labels, and how the appearance of some output label may affect the appearance probability of some input feature, etc. Such reasoning contributes to a global explainable AI method, helping the users to understand how the DNN behaves when processing inputs.

Second, we can use the BN as the basis for a simple monitor, which can be used to identify cases where the classification of the DNN is unreliable.
In particular, we want to identify two situations where the classification of the DNN becomes unreliable: outliers (data that has hardly ever, or not at all, occurred during training), and a significantly different long-term behaviour compared to the training data-based abstraction.
Such a significant {\bf covariate shift} suggests that the network is used in an environment, which is significantly different from its training environment, and therefore its decision making might not be reliable. An {\bf outlier} suggests that the decision making from the network might not be reliable. Both cases may {\bf compromise the safe application of the DNN}.

Third, we design a few coverage metrics over the BN, together with a novel concolic testing method -- based on symbolic execution -- considering the information from both the network and the BN. These metrics and methods improve over earlier structural coverage metrics \cite{sun2018testing} and concolic testing method  \cite{sun2018concolic} for DNNs.
Since the abstract model can be significantly smaller than the original DNN, the effectiveness of the symbolic execution technique for generating relevant new test cases for the DNN is significantly improved.

In addition to the improved scalability, we show that this approach is closer to the design intent for deep learning.
In convolutional neural networks, while neurons are the minimal (syntactic) units, the features on either the input or the hidden layers are the minimal \emph{\textbf{semantic}} units. As such, the main aim of both the DNN architecture, and the back-propagation learning algorithm, is to improve the \emph{\LFs} learned by hidden layers.
There also exist significant research directions, such as interpretability, that aim at understanding what has been learned by a DNN, \eg by visualising the hidden features.
We therefore base our work on the key hypothesis of deep learning that \emph{\textbf{hidden features should be first-class citizens}}, rather than neurons, who do not represent a clear semantics \perse.

The organisation of the paper is as follows.
In the next section, we will present preliminaries about deep neural networks, dimensionality reduction, and Bayesian Networks.
In Section~\ref{sec:abstraction}, we will turn to the presentation of our Bayesian Network-based approach, including a generic framework on extracting \LFs for given layers (Section~\ref{sec:featureextraction}) and feature space discretisation (Section~\ref{sec:discretisation}), concrete strategies for feature extraction and discretisation (Section~\ref{sec:discr-strat}), and the Bayesian network construction (Section~\ref{sec:constr-bayes-netw}).
We then turn in Section~\ref{sec:expl-bayes-netw} to our approach for exploiting the Bayesian Network for the detection of unanticipated operational data.
Our Bayesian Network-based approach is further utilised in Section~\ref{sec:bn-based-coverage} for defining Bayesian Network-based \LF coverage metrics.
We discuss in Section~\ref{sec:cover-guid-test} our adaptation of a concolic testing algorithm to achieve the latter coverage, and present in Section~\ref{sec:implem-n-evaluations} our implementation and evaluation. Finally, we will discuss related works in Section~\ref{sec:related} and conclude in Section~\ref{sec:conclusions}.

\section{Preliminaries}
\label{sec:preliminaries}

\paragraph{Deep Neural Networks}

Let $\network$ be a deep neural network (DNN) of a given architecture.
For a learning model, we use $(X,Y)$ to denote the training data, where $X$ is a vector of inputs and $Y$ is a corresponding vector of outputs such that $|X|=|Y|$.
Let $\inputdomain$ be the input domain and $\outputdomain$ be the set of labels\footnote{Throughout this paper, we will use double-struck capitals to denote domains.}.
Hence, 
$X\subset \inputdomain$.
We may use $x$ and $y$ to range over $\inputdomain$ and $\outputdomain$, respectively.

A network $\network:\inputdomain\rightarrow \dist(\outputdomain)$ can be seen as a mapping from $\inputdomain$ to probabilistic distributions over $\outputdomain$. This means that $\network(x)$ is a probabilistic distribution, which associates each possible label $y\in \outputdomain$ with a probability value (or confidence level) $(\network(x))_y$. We consequently let $f_\network:\inputdomain\rightarrow \outputdomain$ and \(c_{\N}:\inputdomain → [0,1]\) be such that, for any $x\in \inputdomain$,
\[
  f_\network(x) = \argmax_{y\in \outputdomain}\{(\network(x))_y\}%
\mbox.
\]
$f_\network(x)$ returns the label with the greatest probability, \ie
the classification label.

In this work, we consider \emph{sequential} networks: a network $\network$ consists of a sequence of layers $\Layers = （\layer 1,…, \layer K）$, where every layer $\layer i \in \Layers$ contains a set of \(|\layer i|\) \emph{neurons} \(\dimension_i = ｛n_{i,1},…,n_{i,|\layer i|}｝\).
Each neuron \(n_{i,j}\) in \(h_i\) first computes a value \(\hat n_{i,j}\) in some domain that is usually assumed to be the set of Real numbers \RR, and this value is typically computed as a function of the outputs of neurons in layer \layer{i-1}.
The particular semantics of the latter function depends on the role of the layer: this function often consists of a \emph{linear combination} of (a portion of) its inputs
, which means that one has
\begin{equation}
  \label{eq:linear-neuron-function}
  \hat n_{i,j} = W_{i,j}（n_{i-1,1}, …, n_{i-1,|\layer{i-1}|}） + b_i
\end{equation}
where \(W_{i,j}\) and \(b_i\) are coefficients of \emph{weight} and \emph{bias} matrices for the layer, that are the parameters learned by the network during the training phase.

Other functions are also useful, notably to reduce over\-fitting by decreasing the capacity of the network (\ie the total number of weight and bias parameters that need to be learned).
This is for instance the case of \emph{max-pooling} layers, that down-sample their inputs by selecting the maximum over a given subset%
:
\begin{equation}
  \label{eq:maxpooling-neuron-function}
  \hat n_{i,j} = \max
  {s_j（n_{i-1,1}, …, n_{i-1,|\layer{i-1}|}）}
\end{equation}
where each \(s_j\), for \(j∈｛1,…|\layer i|｝\), is a function that gives a subset of all given neuron outputs from the preceding layer (the successive \(j\)'s typically result in a sliding window over all inputs).
The functional semantics of such a layer is therefore non-linear.

Each value \(\hat n_{i,j}\) of a layer \layer i may additionally pass through an \emph{activation function} \(σ_i\) that enables \N to capture non-linearities: the output of neuron \(n_{i,j}\) is called its \emph{activation}, and is then computed as
\begin{equation}
  \label{eq:sigma-activation-function}
  n_{i,j} = σ_i(\hat n_{i,j})\mbox.
\end{equation}
Notable examples of activation functions include \eg ReLU, Sigmoid, tanh, etc.
In this work, we shall focus on the former, which can be defined as:
\begin{equation}
  \label{eq:relu-definition}
  \mathrm{ReLU}(\hat n) ≝ \max ｛0, \hat n｝
\end{equation}

The product of domain of all neuron values in \(h_i\) builds up the \emph{valuation space} \layerdom i of all neurons for layer \layer i (the \(\hat n_{i.j}\)'s); one typically has, therefore, \(\layerdom i = \RR^{|\layer i|}\).
We pose that the valuation of every neuron in the \emph{input layer} \layer 1 corresponds to each component (\eg pixel) of the input \(x ∈ \inputdomain\), so that \(\layerdom 1 = \inputdomain\).
We also denote with \(\hat h_i(x) ∈ \layerdom i\) the set of all \emph{neuron values} 
in layer \layer i (before being passed through any subsequent activation function), when \N is fed with input \(x\).

\figurename~\ref{fig:diagram} presents a simple four layer network, where each of the first three layers contains 4 neurons and the last layer contains 2 neurons.

\paragraph{Example DNNs for Handwritten Digits Recognition}

\begin{table}
  \centering
  \smaller
  \begin{tabular}{clrr}\hline
    Layer & Function specification                & Output shape  & \#weights + \#bias parameters \\ \hline
    \layer 0 & \texttt{conv2d} (convolutional)    &   26 × 26 × 8 & 80 \\
    \layer 1 & \texttt{flatten} (flat)            &   5408        & 0 \\
    \layer 2 & \texttt{dense} (dense)             &   42          & 227\,178 \\
    \layer 3 & \texttt{dense\_1} (dense)          &   10          & 430 \\
    \hline
  \end{tabular}
  \caption{Structure of the DNN \Nms, dedicated to the MNIST dataset; layers \layer 0 and \layer 2 incorporate ReLU activation functions, whereas the output layer \layer 3 involves a classical Softmax to output the predicted label.}
  \label{tab:dnn-Nms}
\end{table}
\begin{table}
  \centering
  \smaller
  \begin{tabular}{clrr} \hline
    Layer & Function specification                & Output shape  & \#weights + \#bias parameters \\ \hline
    \layer 0 & \texttt{conv2d} (convolutional)    &   26 × 26 × 8 & 80 \\
    \layer 1 & \texttt{max\_pooling2d} (max-pooling) &   13 × 13 × 8        & 0 \\
    \layer 1 & \texttt{flatten} (flat)            &   1352        & 0 \\
    \layer 2 & \texttt{dense} (dense)             &   42          & 56\,826 \\
    \layer 3 & \texttt{dense\_1} (dense)          &   10          & 430 \\ \hline
  \end{tabular}
  \caption{Structure of the DNN \Nmx, similar to \Nms except for it incorporates a max-pooling layer.}
  \label{tab:dnn-Nmx}
\end{table}
For the purpose of keeping our illustrations simple, we base most of our examples on a small DNN \Nms, whose structure is summarised in \tablename~\ref{tab:dnn-Nms}.
We have trained \Nms on the MNIST dataset, that consists of labelled black-and-white images of 28×28 pixels, until it reached a validation accuracy of about 98\%.
We have similarly constructed and trained a DNN \Nmx that incorporates a max-pooling layer, summarised in \tablename~\ref{tab:dnn-Nmx}, to further investigate the effect of such filters on both the hidden space and the ability of our test generation algorithm to capture their intrinsically non-linear functional semantics.

\paragraph{Dimensionality Reduction via Feature Extraction}
\label{sec:dimens-reduct-via}

The goal of \emph{dimensionality reduction} techniques consists in computing a mapping from a high-dimensional space into some space of (much) lower dimensionality, called the \emph{feature space}.
Computing approaches for such mappings usually rely on statistical principles and operate on a given sample of high-dimensional data; prominent examples include \emph{Principal Component Analysis} (PCA), \emph{\mbox{\(t\)-}distributed Stochastic Neighbour Embedding} (t-SNE)~\citep{maaten2008visualizing}, or other forms of \emph{kernel tricks}~\citep{theodoridis2003pattern}.
Other techniques that do not specifically address dimensionality reduction \perse, still compute relevant transformations in our context---such is the case of \emph{Independent Component Analysis} (ICA)~\citep{HYVARINEN2000411}.
Each one of these techniques results in mappings that satisfy various properties of interest, among which we notice:
\begin{description}[nosep]
  \newcommand\WDom{\ensuremath{\mathbb{D}}\xspace}%
\item[Linearity:] this property is satisfied when the obtained mapping consists of a pair of matrices \(W ∈ \WDom^{H×L}\)
  and \(B ∈ \WDom^L\) for some field \WDom, such that any element \(x ∈ \WDom^{
    H}\) from the \(H\)-dimensional space is mapped onto the element \(xW + B ∈ \WDom^{
    L}\) in the \(L\)-dimensional feature space.
  Both PCA and ICA belong to the set of techniques that give such linear mappings.
\item[Orthogonality:] PCA computes features one by one, in an effort to minimise the variance of the data.
  The importance of the features -- in terms of their effectiveness in the reduction of variance -- is gradually decreased, although the features are orthogonal (linear independence) to each other.
  Conversely, ICA computes features at the same time, and the generated features may not be orthogonal to each other.
\end{description}
Many available dimensionality reduction techniques compute mappings that are non-linear, such as t-SNE for instance, which is a supervised learning procedure that tries to separate clusters of samples of the same label for adequate visual representations in 2D or 3D spaces.
Whereas some of these techniques are adequate for constructing our BN abstraction, we focus in this work on feature mappings that are \emph{linear}.
This restriction will actually enable us to provide an effective feature-based test-case generation algorithm.

We will consider feature mappings in decomposed form: a mapping will be given as a set \(Λ = ｛λ_j｝_{j∈｛1,…,|Λ|｝}\), where each \(λ_j\) maps the high-dimensional space into the \mbox{\(j\)th-}component of the feature space.
We will use a double-struck capital \FF to denote such a feature space, and indexed \FF's to denote the domain of each one of its individual components (typically \RR).

\paragraph{Bayesian Network}

A \emph{Bayesian Network} (BN) $\B = (V, E, P)$ is a directed acyclic graph (DAG) whose nodes $V$ represent variables in the Bayesian sense~\citep{PGM2009}, \(E\) are edges, and \(P\) maps each node in \(V\) to a set of \emph{probability tables}.
\B is such that each edge in $E$ represents conditional dependencies, and nodes that are not connected (no path connects one node to another) represent variables that are conditionally independent of each other.
In our BN-based analysis, every node that has at least one incoming edge will be attached with a \emph{conditional probability table}, and each source node will be associated with a \emph{marginal probability table}.

\begin{figure}
  \centering
  \includegraphics[width=\textwidth]{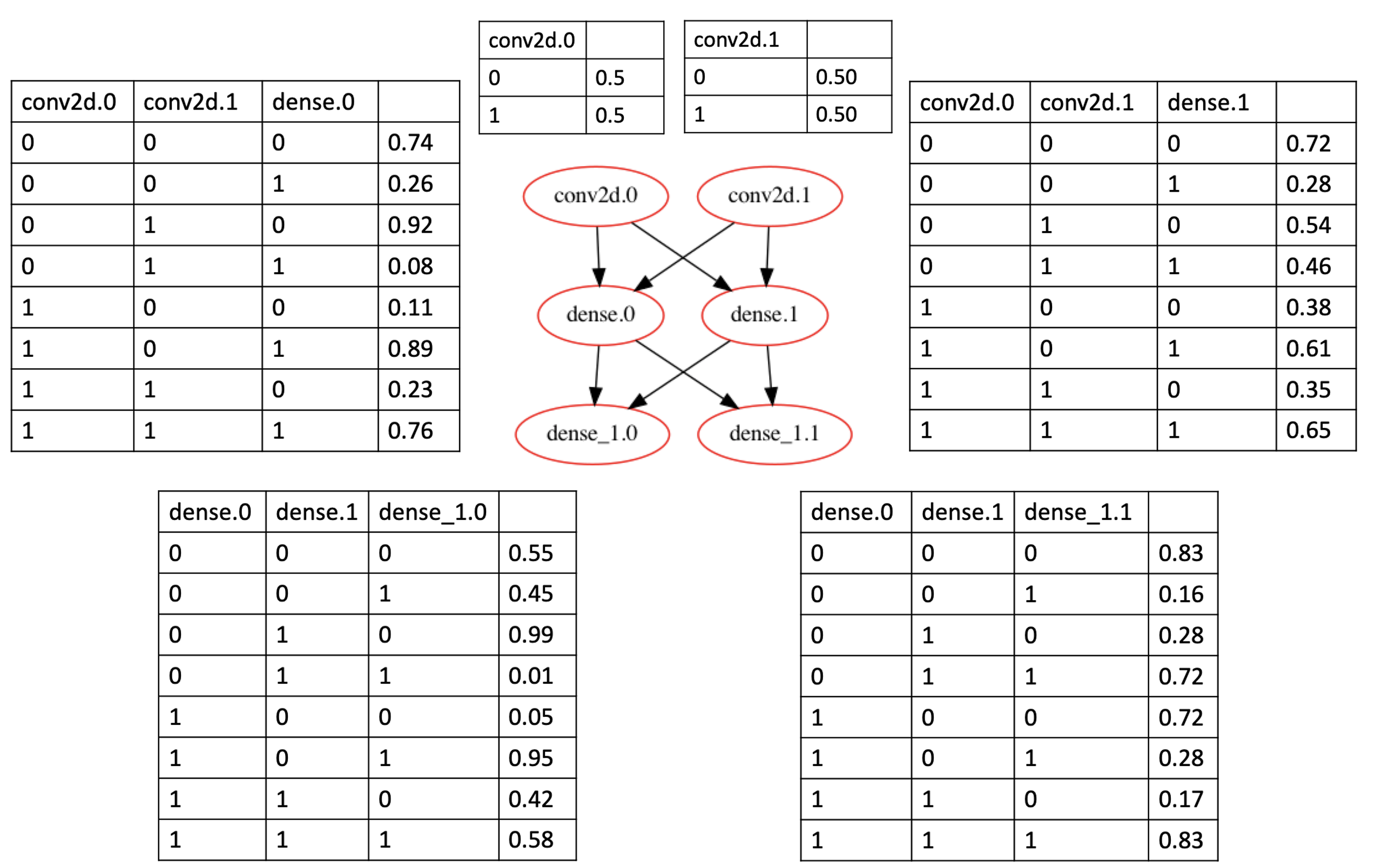}
  \caption{An example Bayesian network abstracted from a network of three hidden layers (one convolutional followed by two fully-connected layers) trained on MNIST dataset.}
  \label{fig:BN}
\end{figure}
\figurename~\ref{fig:BN} presents a simple BN abstracted from \Nms according to the process described in the next Section.
It represents 6 binary variables (that take their values in the set \(｛0,1｝\)), and it thus has 6 nodes, each of which is attached with a probability table.
The two nodes \textsf{conv2d.0} and \textsf{conv2d.1} have no predecessor, and they are thus only associated with marginal probability tables.
From the first table, we can learn that the variable \textsf{conv2d.0} takes value 0 with probability \num{0.5}.
Every other table in this \figurename~represents a set of conditional probabilities: for instance, under the condition that variables \textsf{dense.0} and \textsf{dense.1} respectively hold values 1 and 0, then the variable \textsf{dense\_1.0} takes value 0 with probability \num{0.05}.

\section{Bayesian Network-based Abstraction of DNN}
\label{sec:abstraction}

Let us now turn to the description of our approach for constructing a Bayesian Network in a way that suits out needs for capturing the distribution of neuron valuations in terms of \LFs encoded in each DNN layers, as well as their causal relationships.

\subsection{Extraction of Hidden Features}\label{sec:featureextraction}

Let us assume that some feature extraction technique from Section~\ref{sec:dimens-reduct-via} has been used to analyse the set of neuron values \(\hat n_{i,1},…,\hat n_{i,|\layer i|}\) that are induced by all inputs in a given \emph{training set} \(\Xtrain ⊂ \inputdomain\) at a given layer \(\layer i ∈ \Layers\).
This produces a set of feature mappings $\Features i = ｛\lambda_{i,j}｝_{j ∈ ｛1,…,\lowdim i｝}$, where each \(λ_{i,j}: \layerdom i → \featcomp i j\) maps the neuron valuation space \layerdom i into the \mbox{\(j\)-}th component of the feature space \featdom i for layer \layer i.
The latter space is the product \(\featdom i ≝ \prod_{j∈｛1,…,\lowdim i｝}{\featcomp i j}\mbox.\)

Further, \Features i is such that the neuron values $\hat h_i(x)$ -- \ie before being fed to the activation function \(σ_i\) -- for any input \(x ∈ \inputdomain\), can be transformed into a \mbox{\(\lowdim i\)-}dimensional vector
\begin{equation}
  \langle\lambda_{i,1}∘\hat h_i(x),…,\lambda_{i,\lowdim i}∘\hat h_i(x)\rangle ∈ \featdom i
\end{equation}
where $\lambda_{i,j}∘\hat h_i(x)$ represents the \mbox{\(j\)-}th component of the value obtained after mapping $\hat h_i(x)$ into the feature space.
We will refer to the projection \(λ_{i, j}∘\hat h_i(x)\) as the \emph{\LF valuation induced by \(x\) on component \featcomp i j}.

\begin{leaveout}
  In the context of a layer \layer i where values output by all neurons result from the application of a Rectified Linear Unit (ReLU), we will further identify the projection of the \emph{activation threshold} into the feature space as \(λ_{i}(0) ≝ 〈λ_{i,1}(0), …, λ_{i,t_i}(0)〉\mbox.\)
\end{leaveout}

\begin{figure}[t!]
    \centering
    \includegraphics[width=\textwidth]{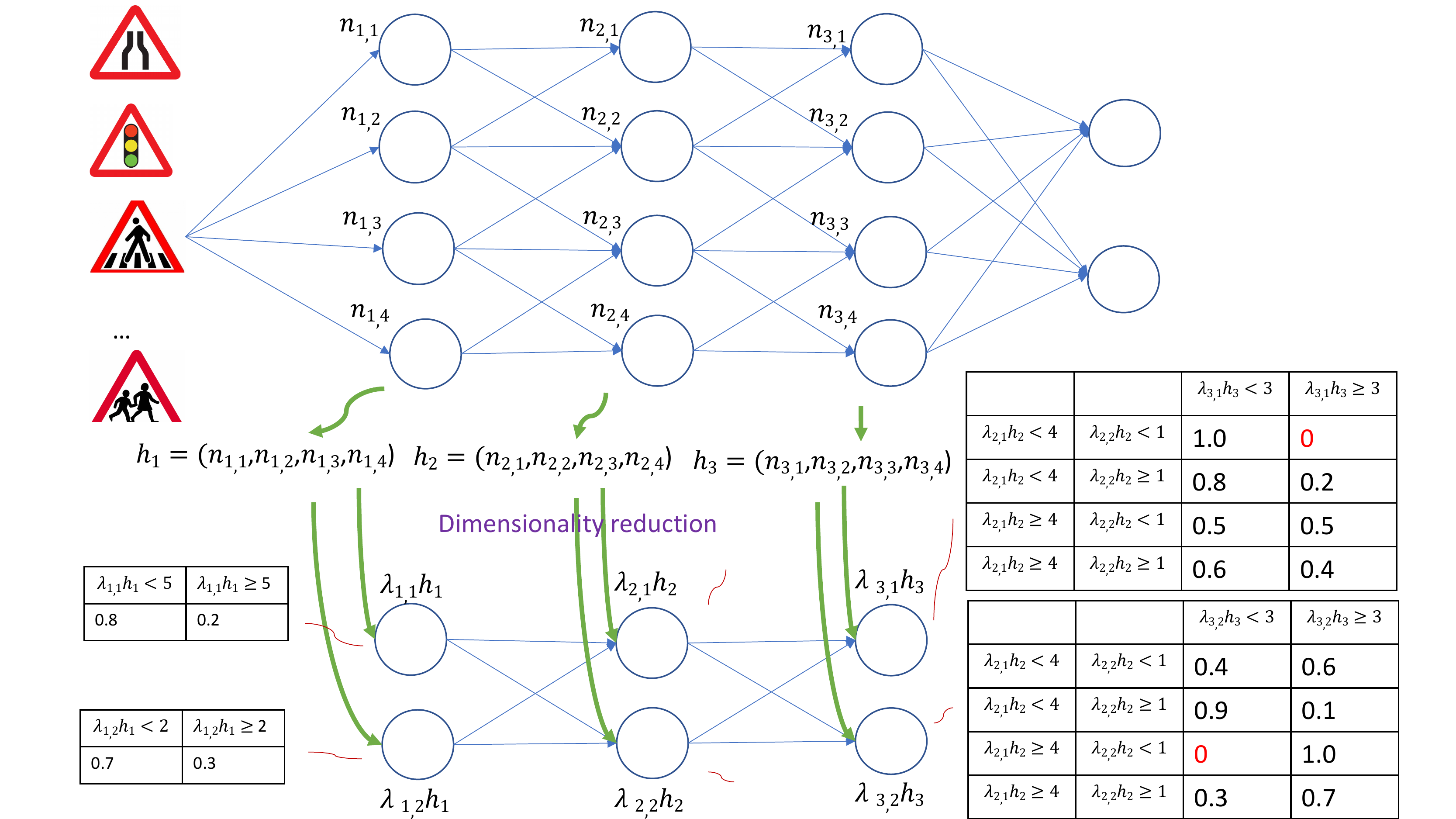}
    \caption{Illustration of our Reduction of Neural Networks to Bayesian Networks}
    \label{fig:diagram}
\end{figure}
Figure~\ref{fig:diagram} gives an illustrative diagram of reducing $\dimension_1, \dimension_2,\dimension_3$ to features.
In particular, each $\dimension_i$ is reduced to two features $\lambda_{i,1}∘\dimension_i$ and $\lambda_{i,2}∘\dimension_i$.

\subsection{Discretisation of the Hidden Feature Space}\label{sec:discretisation}

The feature extraction techniques mentioned in Section~\ref{sec:dimens-reduct-via} result in mappings $\lambda_{i,j}
$ that range over a continuous and potentially infinite domain, such as ℝ.
Yet, our BN-based abstraction technique relies on the construction of \emph{probability tables}, where each entry associates a set of \emph{distinct} \LF values with a probability.
For this construction to be relevant, we therefore \emph{discretise} each \LF component into a \emph{finite} set of sub-spaces.
We first introduce the general process and notations we use for discretising \LFs, and elaborate on various strategies in the next Section.

\paragraph{Discretisation Process}

We define the discretisation process in a similar way as the feature extraction above, and rely on a set of \emph{training inputs} \Xtrain to gather statistics about the projections onto each \LF component of the resulting neuron values.
Given such a component \FF, we use these aggregate statistics to find a \emph{partition} of \FF.

\begin{figure}
  \centering%
  {\input{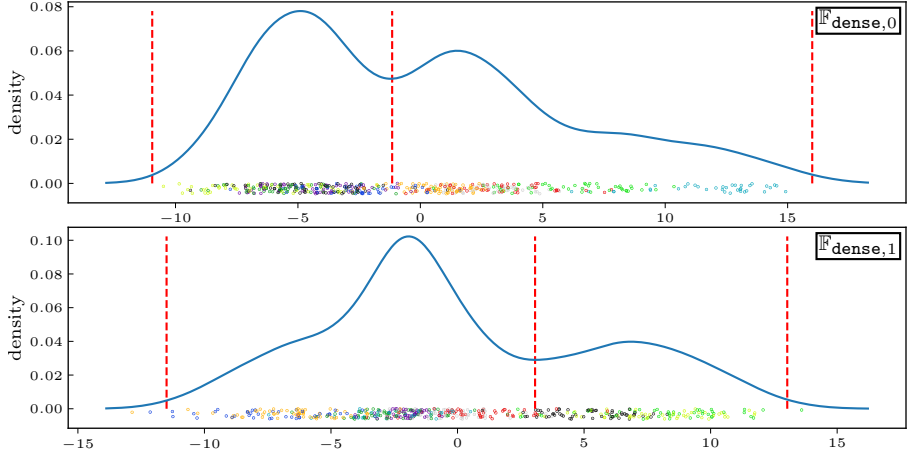}}%
  \caption{
    Projection onto two \LF components \featcomp{\texttt{dense}} 0 and
    \featcomp{\texttt{dense}} 1 of neuron values induced by a sample
    of training data \Xtrain%
    , associated density estimates (solid lines), and interval boundaries for discretisation (dashed vertical lines).
    Although our current strategies for discretisation ignore input labels, we still represent them by colouring the respective dots.}
  \label{fig:mnist_small-dense-dens}%
\end{figure}
We give in \figurename~\ref{fig:mnist_small-dense-dens} an illustration of what the discretisation process produces.
Each plot focuses on one \LF component \featcomp{\texttt{dense}} j of a layer \texttt{dense}, and the clouds of dots represents the respective distribution of hidden feature values \(λ_{\texttt{dense},j}∘\hat h_{\texttt{dense}}(x)\), for \(x\) uniformly drawn from a subset of \Xtrain.
The goal of the discretisation process is therefore to partition each horizontal axis into a set of distinct regions.
In this work, we only consider partitions into finite sets of intervals
.
We describe the remaining elements of \figurename~\ref{fig:mnist_small-dense-dens} in the next Section.

More specifically, the discretisation process produces two objects for each mapping 
\(λ_{i,j}\)
:
\begin{itemize}[nosep]
\item a \emph{finite set of intervals} that partitions the feature component;
\item a total \emph{discretisation function} that maps each neuron valuation for this layer, \ie that belongs to \layerdom i, onto the corresponding interval.
\end{itemize}

\paragraph{Further Notations}

We use \(^{♯}\) exponents to denote discrete spaces, or entities that belong to a discrete space.
Formally, the process of discretising a feature mapping \(λ_{i,j}\) consists in finding a finite set of \(m\) 
right-open intervals \(\Subfeatspaces i j = ｛\subfeatspace i j 1, …
,\subfeatspace i j m｝\) that partitions the codomain \featcomp i j of \(λ_{i,j}\).
\begin{example}[{\LF[L]} partitioning]\label{example:discretisation1}
  We may partition $\featcomp i j = (- \infty, \infty)$ into two intervals \(\Subfeatspaces i j = ｛\subfeatspace i j 1,\subfeatspace i j 2｝\) such that $ \subfeatspace i j 1 $ includes all real numbers strictly lower than 5, and $\subfeatspace i j 2$ includes all real numbers greater or equal than 5.
\end{example}

Given an interval \subfeatspace i j k, we denote with \subfeatlb i j k (resp. \subfeatub i j k) the least element in \subfeatspace i j k (resp. the least element \emph{not} in \subfeatspace i j k).
This discretisation being a partition, one always has \(\subfeatlb i j 1 = -∞\) and \(\subfeatub i j n = ∞\) if the component ranges over \RR.

\begin{example}[Continuing Example~\ref{example:discretisation1}]
We can write $ \subfeatspace i j 1 $ as $(-∞, 5[$ and $\subfeatspace i j 2$ as $ [5, +∞)$.
\end{example}

To obtain the \LF interval that results from all neuron values at layer \layer i, we define the discretisation function \(\featsubsp i j: \layerdom i → \Subfeatspaces i j\) as the total mapping that gives the interval for the \mbox{\(j\)th} component at layer \layer i, which corresponds to the interval obtained for the discretised \LF component \(\featcomp i j\); \ie \(\featsubsp i j ∘\hat h_i(x) = \subfeatspace i j k\) iff \(λ_{i,j}∘\hat h_i(x) ∈ \subfeatspace i j k\).

At last, we define the full \emph{discretised \LF space} for a layer \layer i as the product \(\DiscrFeatsSpace i ≝ ∏_{j∈｛1,…,\lowdim i｝}{\Subfeatspaces i j}\), and use \DiscrFeatsElt i to denote elements of \DiscrFeatsSpace i.

\paragraph{Illustrations}

\begin{figure}
  \centering%
  \input{diagrams/feat-space-discr.tikz}%
  \caption{Schematic representation of the discretised feature space for a single layer \layer i, and two extracted features.
    The ellipse represents the high-dimensional \emph{valuation space} for \layer i, \ie \(\protect\layerdom i\), and we illustrate the discretisation of two feature components \featcomp i 1 and \featcomp i 1 that are induced by mappings \(λ_{i,1}\) and \(λ_{i,2}\), respectively.
    We arbitrarily partition feature \(λ_{i,1}\) into a set of four intervals \(\Subfeatspaces i 1 = ｛\subfeatspace i 1 1, …, \subfeatspace i 1 4｝\), and \(λ_{i,2}\) into three intervals \(\Subfeatspaces i 2 = ｛\subfeatspace i 2 1, …, \subfeatspace i 2 3｝\).
    The application of \(λ_{i,1}\) and \(λ_{i,2}\) on the valuation \(\hat h_i(x)\) induced by some input \(x ∈ \protect\inputdomain\) gives feature values \(λ_{i,1}∘\hat h_i(x)\) and \(λ_{i,2}∘\hat h_i(x)\) that belong to intervals \subfeatspace i 1 2 and \subfeatspace i 2 3, respectively.}
  \label{fig:feat-space-discr}
\end{figure}
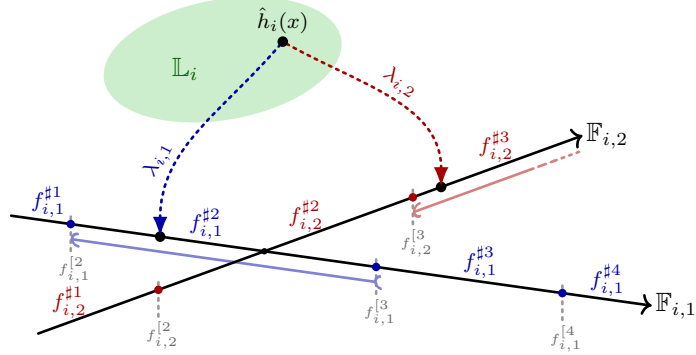
We illustrate in \figurename~\ref{fig:feat-space-discr} the notations that we use to denote the discretisation for a single layer according to some feature mapping \(Λ_i ⊇ ｛λ_{i,1}, λ_{i,2}｝\) and arbitrary discretisations to four and three intervals, respectively: \ie \(\Subfeatspaces i 1 = ｛\subfeatspace i 1 1, …, \subfeatspace i 1 4｝\) and \(\Subfeatspaces i 2 = ｛\subfeatspace i 2 1, …, \subfeatspace i 2 3｝\).
We also show the respective intervals obtained from the neuron values induced by some input \(x\) as \(λ_{i,1}∘\hat h_i(x) ∈ \subfeatspace i 1 2\) and \(λ_{1,2}∘\hat h_i(x) ∈ \subfeatspace i 2 3\); in other words: the application of the feature mapping \(λ_{i,1}\) to the neuron values \(\hat h_i(x)\) at layer \layer i obtained with input \(x\), belongs to the second interval \subfeatspace i 1 2 in the discretisation of \featcomp i 1.
We apply the same principle and obtain the third interval \subfeatspace i 2 3 for \featcomp i 2.

\begin{example}
  We further give in \figurename~\ref{fig:diagram} a more concrete example, where the feature \(\lambda_{1,1}\) is split into the set of ranges \(｛(-∞, 5[,~[5, +∞)｝\mbox,\) and the feature component \(λ_{2,2}\) is partitioned into \(｛(-∞,1[, ~[1, +∞)｝\).
\end{example}

\subsection{Strategies for Hidden Feature Extraction \& Discretisation}
\label{sec:discr-strat}

Predefined strategies need to be employed for both extracting \LFs (\ie to identify a number of mappings that correspond to relevant \LFs), and constructing a discretisation as defined above, for each of them.

At this stage, we can anticipate that the choice of such strategies has a great impact on various properties of the resulting discretised \LF space, such as its ability to accurately separate the hidden feature valuations that are induced by ``sufficiently'' dissimilar inputs, or to isolate hidden sub-spaces that correspond to corner cases.
In particular, we hypothesise that various characteristics of each layer within the neural network, such as its function and position (depth), need to be taken into consideration when selecting discretisation strategies.

Furthermore, the fact that we use the training data to perform extractions and discretisations has an impact on the semantics of the resulting hidden sub-spaces.
Indeed, in doing so we rely on the training data and the set of neuron valuations it induces within the DNN to infer an \emph{abstract representation of its learned behaviours}.
For now, we devise and explain several such strategies
.

\paragraph{Density-based Discretisation}

Let us now refer to the density estimates given in \figurename~\ref{fig:mnist_small-dense-dens}.
These estimations were obtained using Kernel Density Estimation (with Gaussian kernels), which is a standard technique for empirically approximating a probability distribution over a given domain (here, each \(\featcomp{\texttt{dense}} j\)).
We can first observe that the distributions of \LF values exhibit some irregularities that can be exploited to identify relevant intervals.

We adapt DeepGauge's idea to identify corner-cases in test data (\wrt training) among sets of neuron activations by singling out regions of neuron values that are never encountered in test data~\citep{ma2018deepgauge}.
In our case, this naturally translates into the identification of \LF component intervals that are not exercised (enough) by the neuron values induced by the training dataset.

For each \LF component, our strategy for computing intervals using density estimates consists of a simple analysis of the associated density distribution.
We define as interval boundaries the \LF values:
\begin{enumerate*}[(i)]
\item where the density crosses a given lower-bound threshold \(d_{\mathit{min}}\);
\item prominent local minima with density strictly greater than \(d_{\mathit{min}}\).
\end{enumerate*}
In \figurename~\ref{fig:mnist_small-dense-dens}, the dashed vertical lines illustrate the result of such a discretisation strategy for the two considered \LF components.

\paragraph{Further, Simpler Discretisations}

For the purposes of our initial approach, we also investigate two basic discretisation strategies that operate by partitioning each segment \(［\min（λ_{i,j}∘\hat h_i(\Xtrain)）,\max（λ_{i,j}∘\hat h_i(\Xtrain)）］\):
\begin{description}
\item[\(k\)-bins-uniform:] The set of intervals partitions the segment into a given strictly positive number \(k\) of \emph{bins}, all of the same width;
\item[\(k\)-bins-quantile:] The set of intervals is created as above, except that their respective width is calculated so that every interval holds a similar amount of individual projections from the training dataset \(λ_{i,j}∘\hat h_i(\Xtrain)\).
  For instance, the \textit{\(4\)-bins-quantile} strategy proceeds by gathering all projected activations into 4 equally populated bins, and then computes 4 intervals based on the edges of each bin.
\end{description}
The resulting set of intervals is straightforwardly augmented with two left- and right-open intervals to obtain a partition of the \LF component line.
These latter intervals are regions of neuron valuations that we assume correspond to corner cases for the DNN under investigation.

The combined set of extracted and discretised \LFs provides us with a \emph{discrete} and \emph{low-dimensional space} that we can use to reason about the neuron valuations induced by any given input \(x\).
This space will, in particular, allow us to associate each combination of \LF intervals with a measure of its occurrence within a given dataset (other that \Xtrain), as well as reason about the dependencies between intervals of successive layers.

For the sake of conciseness, we will say that such an input (or set of inputs) \emph{elicits} or \emph{exercises} a given \LF interval \subfeatspace i j k, noted \(x ⤳ \subfeatspace i j k\) (or \(X ⤳ \subfeatspace i j k\)), when \(\featsubsp i j∘\hat h_i(x) = \subfeatspace i j k\).
We generalise this notation to a given combination of intervals \(\DiscrFeatsElt i = （\subfeatspace i 1 {k_1}, …, \subfeatspace i {\lowdim i}{k_{\lowdim i}}） ∈ \DiscrFeatsSpace i\) for all \LFs extracted for a layer \layer i as \( x ⤳ \DiscrFeatsElt i ≝ ⋀_{j∈｛1,…,\lowdim i｝}{x ⤳ \subfeatspace i j {k_j}}\mbox.
\)

\subsection{Construction of the Bayesian Network Abstraction}
\label{sec:constr-bayes-netw}

The abstraction that we construct primarily represents the \emph{probabilistic distribution} of the set of \LF values induced by a test sample \(X\).
In other words, given an input \(x ∈ X\), the abstraction allows us to estimate the probability that \(x\) induces a given combination of values for the \LFs that have been learned by the DNN.

Thanks to the layered and a\-cyclic nature of the DNNs that we consider, we can directly characterise the \emph{causal relationship} between the sets of neuron values in various layers \wrt a series of inputs as well.
In other words, given an input \(x ∈ X\), one can in principle estimate the conditional probability of each neuron value at layer \layer i \wrt the probability of every combination of neuron values at layer \layer{i-1}.
By lifting the above relationship from individual neuron values to \LF intervals, we seek to capture \emph{causal semantic relations} that link the features at each layer: in a layer \layer i, and with an input \(x\), the \emph{probability} that a \LF valuation belongs to a given interval in the corresponding feature space is \emph{dependent} on probabilities pertained to \LF intervals at layer \layer{i-1}.

\newcommand\BNaStructure[1]{\ensuremath{\B_{#1}}\xspace}%
\newcommand\BNa[2]{\ensuremath{\B_{#1,#2}}\xspace}%
\newcommand\BNode[2]{\ensuremath{⦇f^{♯}_{#1,#2}⦈}\xspace}%
\newcommand\IPr[2]{\ensuremath{\mathcal{IP}\!（{#2}）}\xspace}%
\newcommand\CPr[4]{\ensuremath{\mathcal{CP\!}_{#1}\!（{#3}｜#4）}\xspace}%
\newcommand\MPr[2]{\ensuremath{\mathcal{P}\!_{#1}{（#2）}}\xspace}%

{\sloppy Let us assume that suitable feature-extraction and discretisation strategies have been employed to obtain a set of discretised feature components \(\AllSubFeatSpaces\N = ｛\Subfeatspaces 1 1, \Subfeatspaces 1 2, …, \Subfeatspaces K {\lowdim K}｝\) for a network \N with layers \((\layer 1, …, \layer K)\).
  We first associate each \LF component \featcomp i j with a discrete (\ie categorical) random variable ranging over \Subfeatspaces i j.
  The causal relationship assumption further gives us the \emph{structure} of a Discrete Bayesian Network, where each node is associated with a discretised \LF, and each edge denotes the dependency between a pair of \LFs in successive layers.
  Given this structure and a set of \emph{inputs} \(X ⊂ \inputdomain\), \emph{Bayesian inference} on the sets of feature intervals \(｛\featsubsp i j ∘\hat h_i(X)｜j∈｛1,…,\lowdim i｝｝\), for \(1≤i≤K\), allows us to estimate the probabilities mentioned above, and construct a BN $\BNa\N X ≝ (V_{\N,X},E_{\N,X},P_{\N,X})$, where
  \begin{equation}
    \label{eq:bn-nodes}
    V_{\N,X} ≝ ｛\BNode i j ｜ \Subfeatspaces i j ∈ \AllSubFeatSpaces\N｝
  \end{equation}
  is the set of nodes, \ie one variable for each extracted feature component,
  \begin{equation}
    \label{eq:bn-edges}
    E_{\N,X} ≝ ⋃_{1 < i ≤ K}｛\BNode{i-1} 1, … , \BNode{i-1}{\lowdim{i-1}}｝×｛\BNode i 1, … ,
    \BNode i {\lowdim i}｝
  \end{equation}
  is the set of edges connecting features in neighbouring layers, and
  \begin{equation}
    \label{eq:bn-conditional-probabilities}
    P_{\N,X}\BNode i j ≝ \begin{cases}
      \MPr 1 {\subfeatspace 1 j k} & \text{~if~} i = 1 \\
      \CPr i x {\subfeatspace i j k} {\DiscrFeatsElt{i-1}} & \text{~otherwise}
    \end{cases}
  \end{equation}
  associates each node with either a marginal probability table (
  for \LFs of layer \layer 1) or a conditional probability table
  (
  for hidden or output layers).
  A conditional probability table \(\CPr i x {\subfeatspace i j k} {\DiscrFeatsElt{i-1}}\) is defined for each feature interval \(\subfeatspace i j k ∈ \Subfeatspaces i j\) for layer \layer i, \wrt each combination of feature intervals \(\DiscrFeatsElt{i-1}
  \) for layer \layer{i-1}, as
  \begin{equation}
    \label{eq:node-cond-prob}
    \CPr i x {\subfeatspace i j k} {\DiscrFeatsElt{i-1}} ≝
    \P{%
      x ⤳ \subfeatspace i j k｜%
      x ⤳ \DiscrFeatsElt{i-1}
    }\mbox.
  \end{equation}
  Intuitively, \(\CPr i x {\subfeatspace i j k} {\DiscrFeatsElt{i-1}}\) gives the probability that any input \(x ∈ X\) chosen uniformly 
  exhibits feature interval \(\subfeatspace i j k\), knowing that it exhibits intervals \(\DiscrFeatsElt{i-1}\) for all extracted features at layer \layer{i-1}
  .
  \par}

One can compute \emph{marginal probabilities} for \LF intervals that pertain to hidden and output layers based on the conditional probability tables as defined in Eq~(\ref{eq:node-cond-prob}): the \emph{unconditional probability} obtained for every interval \(\subfeatspace i j k ∈ \Subfeatspaces i j\) is then
\begin{equation}
  \label{eq:marginal-prob}
  \MPr i {\subfeatspace i j k} ≝ \P{
    x ⤳ \subfeatspace i j k}\mbox.
\end{equation}
Eq.~(\ref{eq:marginal-prob}) gives the probability that, given any input \(x\) chosen uniformly in \(X\), \(λ_{i,j}∘\hat h_i(x)\) belongs to a given feature interval \subfeatspace i j k, without any knowledge on the neuron values.
\LFs[H] pertaining to the input later are not subject to any conditional dependence in the BN we construct.
We therefore use \MPr 1 {\subfeatspace 1 j k} in Eq.~(\ref{eq:bn-conditional-probabilities}), which estimates the unconditional probabilities associated with \LF intervals of \layer 1.

Observe that, according to Eq.~(\ref{eq:bn-edges}), the value of each feature extracted from a layer \layer i is assumed to directly depend on the value of every feature for layer \layer{i-1}; this assumption is further reflected in the construction of conditional probability tables in Eq.~(\ref{eq:bn-conditional-probabilities}).
\figurename~\ref{fig:diagram} contains an example BN, where the above assumption induces full connections between the nodes pertaining to successive layers.
\figurename~\ref{fig:diagram} additionally gives two conditional probability tables for $\featsubsp 3 1 ∘ \dimension_3$ and $\featsubsp 3 2 ∘ \dimension_3$, respectively; it also shows two unconditional probability tables for $\featsubsp 1 1 ∘ \dimension_1$ and $\featsubsp 1 2 ∘ \dimension_1$, respectively.
The tables are attached to their respective BN nodes.

\subsection{Preserved Property}

First of all, we show that the constructed BN is an abstraction of the NN.
Given a finite set $X$ of inputs, an abstraction constructs a set $X' \supseteq X$ that generalises $X$ to more elements \cite{DBLP:journals/corr/abs-1911-09032}. Given an input $x$ and a BN $\BNa\N X$, we are able to check the probability of $x$ on $\BNa\N X$, i.e., $\BNa\N X(x)$. We say that $x$ is included in $\BNa\N X$ if $\BNa\N X(x)>0$.
The following lemma suggests that every sample in $X$ is included in $\BNa\N X$:

\begin{lemma}\label{lemma:abstraction}
  All inputs $x$ in the dataset $X$ are included in the $\BNa\N X$ with probability greater than 0.
\end{lemma}

Let $X'$ be the set of inputs that satisfy $\BNa\N X(x)>0$. This lemma suggests that $X'\supset X$. Therefore, $\BNa\N X$ defines an abstraction of the dataset $X$.
This abstraction also suggests that the abstraction assumption -- \ie inputs that are outliers \wrt the abstraction are also outliers \wrt the original neural network -- is reasonable because $X'\supset X$.

\section{Exploiting the Bayesian Network}
\label{sec:expl-bayes-netw}

\subsection{Probabilistic Inference}

Given a BN \BNa\N X, we can conduct probabilistic inference to understand how the original neural network processes inputs. By doing this, we  gain a better understanding about the ``black-box'' neural network -- and improve its interpretability/explainability -- by conducting causal and evidential reasoning.

\paragraph{Causal Reasoning}

is to understand how the up-stream causes may affect the down-stream effects.
Typically, it considers queries such as how the appearance of some input features may statistically affect how a given \LF or output label is exercised.

\newcommand\MAP[1][]{\ensuremath{\mathit{MAP}\lambdaone{#1}}\xspace}
\begin{sloppypar}
  A BN is a joint probability $P(\DiscrFeatsElt{1},…,\DiscrFeatsElt{K}) = \prod_{i=1}^K\prod_{j=1}^{\DiscrFeatsElt{i}} P_{\N,X}\BNode i j$---\cf Eq.~(\ref{eq:bn-conditional-probabilities}).
  If we have evidence on the input feature by having, \eg $\BNode 1 {j_1}=v$ for some value $v$, then we will be able to reason about the maximum a posteriori (\MAP) estimation of the output labels, $\MAP[\BNode K {j_2}｜ \BNode 1 {j_1}=v]$, which represents the maximum probability the output label $j_2$ can have after considering not only the evidence $\BNode 1 {j_1}=v$ but also all possible values of other nodes on the BN.
\end{sloppypar}

\paragraph{Evidential Reasoning} is to understand how the availability of evidence on the down-stream effects may affect the up-stream causes.
Typically, it considers queries such as how the appearance of some output label may affect the appearance probability of some input feature.

Considering a given label $j_2$ (that is, we have an evidence that $\BNode K {j_2}=1$ and $\forall j\neq j_2:\BNode K {j}=0$), the probability of $\MPr 1 {\subfeatspace 1 j k}$ may change to reflect the availability of the evidence.
It is possible that some features may be more important for some specific label.

\subsection{Runtime Monitoring}

We propose two ways of monitoring the behaviour of a DNN through its associated BN.
The first one is to monitor the distribution shift by comparing a BN obtained from operational data and the original BN obtained using the training data.
The second is to monitor the possibility of a given input to determine the risk of a safety violation by checking if it is an outlier.

\paragraph{Monitoring Distribution Shift}

Let $\BNa\N X=(V_{\N,X},E_{\N,X},P_{\N,X})$ be the BN we learned from the training dataset. It can be seen as a generative model for the dataset $X$.
Covariate shift, an important kind of distribution shift, refers to the change in the distribution of the input variables present in the training and the test data.
Now, assume that we have collected a set $X'$ of operational data.
We can use $X'$ to construct another BN $\BNa\N{X'}=(V_{\N,X},E_{\N,X},P_{\N,X'})$ that has the same \emph{structure} as \BNa\N X (\ie the vertices and the edges remain), but where the probability tables are re-generated according to $X'$.

By comparing the probabilities in \BNa\N X and \BNa\N{X'}, we can assess whether some of the probabilistic causal relations are changed.
Indeed, a change on probabilistic causal relation may imply the change on a conditional probability represented by the BN, on which the probabilistic causal relation is a factor.

\paragraph{Detecting Outliers}

Unlike the above methods which only consider the closeness of the new input with the training dataset in terms of geometric distance, we can assess whether any new input is an outlier by considering its probability on the BN: any input \(x'\) for which
\begin{equation}\label{equ:rare}
  \BNa\N X(x') = 0
\end{equation}
will be identified as an outlier.
As implied by Lemma~\ref{lemma:abstraction}, our abstraction is sound because all training instances are considered and none of them will lead to the case of Eq.~(\ref{equ:rare}).

\section{Bayesian Network-based Feature Coverage}
\label{sec:bn-based-coverage}

Given a trained DNN \N, the process described in Section~\ref{sec:constr-bayes-netw} provides a method for constructing a BN \BNa\N X, whose \emph{structure} is derived using assumed dependencies between \LFs that are extracted and discretised by means of the techniques specified in Section~\ref{sec:discr-strat}.
The probabilities in \BNa\N X are further inferred using the set of \LF valuations induced by \(X\), and provide useful insights on whether \(X\):
\begin{enumerate}[(i)]
\item induces a varied range of valuations for each extracted \LF; and
\item exercises the assumed relationships between \LFIs.
\end{enumerate}
Our BN abstraction can therefore be used to define some quality metric for the set \(X\), in terms of the \LFs that are captured in layers of \N that \(X\) does or does not exhibit.
We elaborate on this idea in this Section, and develop new coverage metrics that can be used to assess the quality of test datasets.
We show in the subsequent Section how these metrics can also be exploited to generate new inputs for the purpose of increasing the range of \LF valuations and causal relationships induced by such datasets.

\subsection{Eliciting Semantic Assumptions Missed by Test Data}

Observe that a marginal probability \(\MPr i {\subfeatspace i j k} < ε\) computed based on a BN \BNa\N X, where ε is a small-enough probability, expresses that \(\subfeatspace i j k\) is a \LFI that is rarely elicited by any input in \(X\).
Similarly, a conditional probability \(\CPr i x {\subfeatspace i j k} {\DiscrFeatsElt{i-1}} < ε\) expresses that an assumed causal dependency between intervals for all \LFs of layer \layer{i-1} and a \LFI \subfeatspace i j k for \layer i is not exercised by \(X\).
Under the assumption that the hidden sub-space that underlies the structure of \BNa\N X reflects high-level (semantic) behaviours that \N must capture, then both cases lead to clearly identified \emph{semantic assumptions} that are not tested by any input from \(X\), or not tested enough.
We now turn these observation into more formal definitions of coverage metrics.

\paragraph{Hidden Feature Coverage}

The first BN-based coverage metric that we define disregards conditional probabilities, and concentrates on each individual \LFI in isolation:
\newcommand\BNFCov[2]{\ensuremath{\mathrm{BFCov}（\BNa{#1}{#2}）}\xspace}%
\newcommand\BNFdCov[2]{\ensuremath{\mathrm{BFdCov}（\BNa{#1}{#2}）}\xspace}%
\newcommand\BNFxCov[2]{\ensuremath{\mathrm{BFxCov}（\BNa{#1}{#2}）}\xspace}%
\begin{defn}[BN-based Feature Coverage]
  \label{def:bn-based-feature-coverage}
  Given a trained DNN \N, the \emph{BN-based feature coverage} of a \emph{non-empty} set of inputs \(X ⊂ \inputdomain\) is obtained via the BN abstraction \BNa\N X as
  \begin{equation}
    \label{eq:bn-fc-crit}
    \BNFCov\N X ≝
    \dfrac{1}{\left|\strut V_{\N,X}\right|}
    \sum_{\BNode i j ∈ V_{\N,X}}
    \dfrac{%
      \left|｛\subfeatspace i j k ∈ \Subfeatspaces i j｜ \MPr i {\subfeatspace i j k} ≥ ε｝\right|
    }{%
      \left|{\Subfeatspaces i j}\right|
    }\mbox.
  \end{equation}
\end{defn}
Informally, \(\BNFCov\N X\) ranges over \(]0,1]\), and gives the proportion of features that are exercised (enough) by \(X\)---this measure cannot be null if ε is sufficiently small, since the sum of all the entries of the probability tables is always one.

\begin{example}
  \label{example:hidden-feat-coverage}
  Consider the BN given in \figurename~\ref{fig:diagram}.
  For layer \layer 3, the following marginals can be computed based on the given two conditional probability tables: \(\Pr(λ_{3,1}∘h_3(x) < 3) \approx 0.725\), \(\Pr(λ_{3,1}∘h_3(x)≥3) \approx 0.275\), \(\Pr(λ_{3,2}∘h_3(x) < 3) \approx 0.4\), and \(\Pr(λ_{3,2}∘h_3(x)≥3) \approx 0.6\).
  Assuming a dataset \(X\) was used to infer the probabilities in the figure (and assuming similar non-negligible marginal probabilities for the nodes pertained to layers \layer 1 and \layer 2), then we obtain \(\BNFCov\N X = 1\).
  This means that \(X\) adequately covers the full range of \LFs values learnt by the abstracted DNN.

  Yet, the null conditional probabilities in the figure indicate that some combinations of \LFIs between \layer 2 and \layer 3 are not exercised by \(X\).
\end{example}

\paragraph{Hidden Feature-dependence Coverage}
\label{sec:hidden-feat-dependence-cover}

We further exploit the insights on causal relationships exercised by a dataset \(X\) that the conditional probabilities of a BN \BNa\N X provide to define the following coverage metric:
\begin{defn}[BN-based Feature-dependence Coverage]\sloppy
  \label{def:bn-based-feature-dependence-coverage}
  Given a trained DNN \N, the \emph{BN-based feature-dependence coverage} of a \emph{non-empty} set of inputs \(X ⊂ \inputdomain\) is obtained via the BN abstraction \BNa\N X as
  \begin{equation}
    \label{eq:bn-fdc-crit}
    \BNFdCov\N X ≝
    \dfrac{1}{\left|\strut V^+_{\N,X}\right|}
    \sum_{\BNode i j ∈ V^+_{\N,X}}
    \frac{%
      \left|｛%
        \begin{aligned}
          (\subfeatspace i j k, \DiscrFeatsElt{i-1}) ∈ \\
          \Subfeatspaces i j × \DiscrFeatsSpace{i-1}
        \end{aligned}
        ｜
        \begin{aligned}
          \CPr i x {\subfeatspace i j k} {\DiscrFeatsElt{i-1}} ≥ ε \\
          ∨\quad\MPr i {\subfeatspace i j k} < ε
        \end{aligned}
        ｝\right|
    }{%
      \left|\Subfeatspaces i j × \DiscrFeatsSpace{i-1}\right|
    }
  \end{equation}
  where \(V^+_{\N,X} ≝ ｛\BNode i j ∈ V_{\N,X}｜ i > 1｝\) are all nodes in \BNa\N X that represent \LFs extracted for hidden or output layers.
\end{defn}
\BNFdCov\N X gives the proportion of assumed causal relationships between features of successive layers that are exercised (enough) by \(X\).
Observe that this metric only takes conditional probabilities into account; therefore, the first layer (\ie for \(i = 1\)) is taken out of the summation, and the normalisation factor is adjusted accordingly.
Further, the set above the innermost fraction includes the conditional probability entries pertaining to \LF intervals that are \emph{not} elicited by \(X\) (\ie for which \(\MPr i {\subfeatspace i j k} < ε\)).
We eliminate the impact of feature coverage as measured in Def.~\ref{def:bn-based-feature-coverage} on feature-dependence coverage in this way, \ie so that \BNFdCov\N X only captures the causal relationships between \LFs that are not exercised by \(X\), whatever the actual coverage of \LFs that \(X\) induces.

At last, we straightforwardly combine the two above metrics to provide a consistent coverage measure that is based on every probability entry of the BN:
\begin{equation}
  \label{eq:bfxc}
  \BNFxCov\N X ≝ \BNFCov\N X × \BNFdCov\N X\mbox.
\end{equation}

\begin{example}
  \label{example:hidden-feat-dependence-coverage}
  Continuing Example~\ref{example:hidden-feat-coverage} (and, again, further assuming non-negligible entries in the conditional probability tables that are not shown in \figurename~\ref{fig:diagram}), we obtain
  \[
    \BNFdCov\N X = \frac{1}{4}×（2×\frac{8}{8}+2×\frac{7}{8}） = \frac{15}{16} \approx 94\%\mbox.
  \]
\end{example}

\paragraph{Coverage Criteria}

We trivially derive the following test criteria from the coverage metrics given in Defs.~\ref{def:bn-based-feature-coverage} and~\ref{def:bn-based-feature-dependence-coverage}:

\begin{defn}[BN-based Feature Coverage Criterion]
  \label{def:bfc-crit}
  A non-empty set of inputs \(X ⊂ \inputdomain\) satisfies the \emph{BN-based feature coverage criterion} that is obtained via the BN abstraction \BNa\N X iff \(\BNFCov\N X = 1\).
\end{defn}
\begin{defn}[BN-based Feature-dependence Coverage Criterion]
  \label{def:bfdc-crit}
  A non-empty set of inputs \(X ⊂ \inputdomain\) satisfies the \emph{BN-based feature-dependence coverage criterion} that is obtained via the BN abstraction \BNa\N X iff \(\BNFdCov\N X = 1\).
\end{defn}

\subsection{Advantages of the Abstraction and Coverage Criteria}

In the following, we discuss a few advantages of our BN-based approach to modelling DNN behaviours, and the adjoined coverage criteria.

\paragraph{From Structural Coverage to Semantics Coverage}

Our BN-based abstraction allows us to formulate semantics-based coverage criteria: we are now able to work directly with features instead of neurons.
As argued in the introduction, in machine learning, features are regarded as the basic elements and should be dealt with directly.

\paragraph{Working with Semantics with Minimal Adaptation}
\begin{sloppypar}
  Definitions~\ref{def:bfc-crit} and~\ref{def:bfdc-crit} resemble structural versions of coverage criteria, that are directly based on neuron activations~\citep{PCYJ2017, ma2018deepgauge, sun2018concolic}.
  Actually, $\BNFCov\N X$ can be seen as the BN counterpart of the neuron coverage \cite{PCYJ2017} and $\BNFdCov\N X$ can be seen as the BN counterpart of the MC/DC \cite{sun2018concolic}.
  Therefore, our BN representation of hidden space coverage provides us with a means to enhance existing structural approaches for assessing the quality of trained DNNs with semantics, without requiring drastic changes to the basic principles.
  We will demonstrate this aspect further in the next Section, by showing how the core test case generation algorithm of \DeepConcolic can be adapted to the above criteria.
\end{sloppypar}

\paragraph{Evaluating Suitability of Feature Extraction Techniques on a Dataset}

Our framework is general and can work with any feature extraction technique. Therefore, it can potentially be applied to evaluate and compare the quality of applying feature extraction techniques on some given dataset---it is known that there does not exist a single feature extraction technique that works best on all datasets.
Certainly, to have a fair evaluation and comparison, the dataset needs to be of ``golden standard''.

\paragraph{Improved Scalability}
The size---measured by the number of nodes---of the BN is adjustable, and usually significantly smaller than the original neural network.
However, as hinted in Eq.~(\ref{eq:bn-conditional-probabilities}), the size of the underlying conditional/joint probability tables at a given layer \layer i grows exponentially in the number of nodes used to represent layer \layer{i-1} (\ie its number of \LFs) and in the size of their respective partitions.

Still, if the number of extracted \LFs and their respective partitioning is kept small enough, then the scalability of any analysis method based on our BN representation can be significantly better than an analysis method that is directly based on the neural network's structure.

\paragraph{Connection to Boxing-Clever, and Beyond}
\begin{sloppypar}
  The production of the marginal probability tables of the input layer can partition the input space into a set of hyper-rectangles.
  Each hyper-rectangle is associated with a set of constraints (or half-planes), such that each constraint comes from one \LF of the input layer.
  Therefore, by working with the marginal probability tables, it is easy to work out some functionalities that are enabled in Boxing-Clever~\cite{box-clever}, \eg the computation of \textbf{empty hyper-rectangles}.
  Note that in our case, the hyper-rectangles are based on \LFs instead of neurons.
\end{sloppypar}

Beyond the input layer, every 0-valued entry of a conditional probability table (of some hidden layer) represents a missing training example in satisfying the causal relation represented by the 0-valued entry. By back-propagation of the 0-value to the input layer, it is possible to identify an \textbf{empty region} -- not necessarily an empty hyper-rectangle -- of the input space such that any input in the region satisfies the causal relation represented by the 0-valued entry.

\paragraph{Detection of Unintended Memorisation and its Implication to Safety Assurance}

Intuitively, if the test dataset \(X\) is deemed representative of realistic inputs, then low-probability entries in the Bayesian Network \BNa\N X may reveal out-of-distribution training data samples.
\citet{236216} have suggested that a neural network may unintendedly memorise these out-of-distribution data samples, leading to a negative effect on its generalisation ability~\cite{DBLP:journals/corr/abs-1906-05271}.
We have discussed in~\cite{zhao_safety_2020,zhao_safety_2021} the relation between generalisation ability with the failure rate of neural networks in operation, under a safety assurance framework.

\section{Concolic Approach for Test Case Generation}
\label{sec:cover-guid-test}

We have extended the \DeepConcolic approach for testing DNNs that was pioneered by~\citet{sun2018concolic} with our new test criteria in order to operationally assess the utility of the above constructions and definitions.
Our goal in doing so is to check whether our new definitions for \LF coverage can be used to guide the generation of test cases.

\subsection{\DeepConcolic in a Nutshell}
\label{sec:deepc-nutsh}

The \DeepConcolic approach for test case generation consists of alternating \emph{concrete} executions and \emph{symbolic} analyses of the DNN under test.
The concrete executions actually evaluate the test input using the trained DNN, and report on the level of coverage achieved.
The symbolic analyses, on the other hand, aim at synthesising new test inputs based on some \emph{test target} that is chosen so as to increase the considered coverage metric.
This approach is provably more efficient at constructing sets of test inputs that achieve very good coverage than a basic random generation of test inputs.

\paragraph{Test Case Generation via Symbolic Analysis}

The symbolic analysis is performed on a symbolic model that constrains every neuron activation up to a layer \layer i according to the functional semantics of each respective layer, with the addition of neuron valuation constraints that encode a given test target (such as a change in the sign of a particular neuron output, for instance).
The goal of the analysis is then to solve the problem of finding a value for every input neuron, such that some distance metric \wrt a legitimate test input is minimised, while satisfying all the above constraints.
In the case of input images, typical distance metrics for assessing the similarity between inputs include the Chebyshev distance \(L_∞\), which uses the maximum absolute difference between any one pixel or colour component between two images as a measure of their distance, or the \(L_0\) ``norm'', that simply counts the number of pixels or colour components that differ.
The set of values obtained for every input neuron consists of a new test case that is in principle guaranteed to fulfil the test target; this input is added to the set of generated test cases if some \emph{oracle} mechanism determines that it is ``close enough'' to a reference test input.
The latter requirement serves as a ``plausibility'' filter to only retain realistic inputs.

\paragraph{Linear Programme and Distance Metric}

The problem that is solved at each symbolic analysis step of \DeepConcolic is a Linear Programme (LP), and in our case, the symbolic analysis consists in solving an LP as well.
Such a problem is built as a set of \emph{constraints} \Constr{⋅} that involve continuous (free) \emph{variables}, associated with an \emph{optimisation objective}.
A solution to the problem consists of an assignment for every one of these variables that meets both the set of constraints and the optimisation objective.

The specification of an LP problem requires that every constraint and objective function involved be \emph{linear inequalities} or \emph{linear arithmetic expressions}.
This requirement constrains the distance metric that we can use to compare input images, which must therefore be linear as well.
Among the metrics listed above, only the Chebyshev distance \(L_∞\) satisfies this requirement.
We will therefore assume this distance metric below, and will denote with \LInf x y the distance between two vectors of same dimension \(x\) and \(y\).

\paragraph{Convergence \& Approximations}
\label{sec:convergence}

In some cases, reliance on a linear encoding of functional semantics may induce losses in analysis precision.
This translates in either:
\begin{description}[left=0pt,style=sameline,leftmargin=\parindent]
\item[over-approximation:] incorrect solutions, \ie the LP-problem over-approximates the actual behaviours of the DNN on the given input;
\item[under-approximation:] infeasible problems derived from legitimate behaviours, \ie the LP-problem under-approximates the actual functional behaviours of the DNN on the given input.
\end{description}
In \DeepConcolic, such an imprecision typically arises in layers whose functional semantics cannot be precisely encoded using linear arithmetic and inequalities.
Such is the case for max-pooling layers for instance, that require \adhoc treatments for its encoding as a set of linear constraints.

One can easily check whether the input that is extracted from a solution to the LP problem exhibits the intended behaviours so as to discard inputs that do not meet the desired test target; keeping such inputs may also help better exploring the input-space.
Note, however, that both under- and over-approximations may impair the convergence of the overall generation algorithm.
We elaborate on this particular aspect for our adaptation in the next Section, and investigate it further experimentally in Section~\ref{sec:convergence-aspects}.

\subsection{Concolic Test Generation with BN-based Criteria}
\label{sec:extens-deepc-fram}

In our case, a test target, that we will denote \(\Target{\subfeatspace i j k}\), consists of a (set of) \LF interval(s) that should be elicited by the test input to be generated; such is the case for the criterion given in Def.~\ref{def:bfc-crit} above for instance, for covering a rare \LFI.

\newcommand\Constraints{\ensuremath{\mathit{Constraints}}\xspace}%
\newcommand\Xok{\ensuremath{X^{\mathit{ok}}}\xspace}%
The overall procedure is parameterised with the structure of the Bayesian Network based on which our coverage metrics are defined: \ie we assume here that suitable feature extraction and discretisation have been applied on a training sample \Xtrain.
The computation starts by randomly sampling an initial set of test inputs \(X_0\) that is correctly classified by \N, and initialising the probability tables in the Bayesian Network to produce \BNa\N{X_0}.
Let \(\Xok_0 = X_0\) be the initial set of candidates from which new tests can be derived.
Then, each iteration \(i\) of the test case generation process proceeds according to the following operations:
\label{deepconcolic-lp-analysis-algo}
\begin{enumerate}[ref={step~\arabic*}]
\item\label{step:start}%
  Identify a test target \(t = \Target{⋅}\) that is not yet met by the current set of input test cases \(X_i\).
  This step is performed by means of an analysis of the marginal or conditional probability tables in \BNa\N{X_i};
\item Select a test input \(x ∈ \Xok_i\) according to some heuristics, such as some closeness to the identified target \(t\);
\item\label{step:lp-construction}%
  Construct an LP problem based on \(t\): this problem comprises a set of constraints \Constraints, and an optimisation objective that seeks to minimise the distance between activations of input neurons and \(x\).
  This problem is formulated as:
  \begin{equation}
    \begin{split}
      &\text{Minimise: } \LInf{(n_{1,1},…,n_{1,|\layer 1|})}{(x_{1,1},…,x_{1,|\layer 1|})}\\
      &\text{Subject to: } \Constraints
    \end{split}
  \end{equation}
  where \(n_{1,1},…,n_{1,|\layer 1|}\) is the set of all input neurons in \N;
\item Solve the LP problem, and extract the newly generated test input \(x'\) from values of input neurons: \(x' = (n_{1,1},…,n_{1,|\layer 1|})\);
\item Keep the generated input \(x'\) if it passes the oracle \emph{and, optionally, if it actually improves on \(x\) \wrt the target \(t\)}\label{step:selection4progress}: in such a case, let \(X_{i+1} = X_i ∪ \{x'\}\); otherwise, let \(X_{i+1} = X_i\) and \(\Xok_{i+1} = \Xok_i\), and continue from \ref{step:start};
\item Let \(\Xok_{i+1} = \Xok_i ∪ \{x'\}\) if \(f_\N(x') = f_\N(x)\), \(\Xok_i\) otherwise.
  The latter case indicates that \N does not output the same classification label for \(x'\) than for \(x\), which means that \(x'\) is considered \emph{adversarial} for \N, as \(x'\) is both deemed close enough to \(x\) from which it is derived, and it is not assigned the same label as \(x\) by \N;
\item Update probabilities in \BNa\N{X_i} to account for the new test \(x'\) and construct \BNa\N{X_{i+1}}; the test case generation continues if the test criterion obtained via this new BN is not satisfied.
\end{enumerate}
The three steps from the algorithm above that need to be specialised according to the sought-after BN-based test criteria consist of:
\begin{enumerate}[(i)]
\item the identification of a test target by analysing \BNa\N{X_i};
\item the selection of a test input;
\item the construction of the set of constraints \Constraints.
\end{enumerate}

Test oracle aside, the selection criterion in \ref{step:selection4progress} embodies a practical measure that we employ to counter a consequence of the loss of precision that is induced by dimensionality reduction.
We discuss this aspect further in Section~\ref{sec:helping-convergence} below.
We first elaborate on \ref{step:lp-construction}, which involves an encoding of the functional behaviours of the DNN up to some layer of interest, that is indicated by the test target.

\subsubsection{Encoding the Neural Network}
\label{sec:encod-neur-netw}

Given a test target that aims at obtaining a test input that exhibits a given \LFI at layer \layer k, we first construct a symbolic encoding of the neural network up to \layer k.
We reuse the constraints suggested by \citet{sun2018testing} to construct this encoding of layer behaviours.
Precisely, for each neuron $n_{i,j}$ for $i>1$, and assuming ReLU activation functions, we have two variables $\hat n_{i,j}$ and $n_{i,j}$ and constraints that can directly be derived from Eqs.~(\ref{eq:linear-neuron-function}) and~(\ref{eq:relu-definition}) in Section~\ref{sec:preliminaries}:
\begin{gather*}
  \label{eq:encoding-linear-layers}
  \hat n_{i,j} = W_{i,j}（n_{i-1,1}, …, n_{i-1,|\layer{i-1}|}） + b_i\mbox, \\
  n_{i,j} ≥ \hat n_{i,j}\qquad\mbox{~and~}\qquad n_{i,j} ≥ 0
\end{gather*}
where $W_{i,j}$ and $b_i$ are weight and bias parameters in layer $i$, respectively.

The case of layers that induce non-linear functional behaviours requires an alternative approach, since they cannot be directly encoded in the same way as in Eq.~(\ref{eq:encoding-linear-layers}).
To circumvent this issue, our approach consists in constraining the neurons of these layers in such a way that they exhibit the same behaviour as that induced by the candidate input \(x\).
Considering for instance a max-pooling layer \layer i (for which we gave the functional behaviour in Eq.~(\ref{eq:maxpooling-neuron-function}) in Section~\ref{sec:preliminaries}), the idea is first to capture its selection behaviour using the following binary matrix %
\(S^x ∈ ｛0,1｝^{|\layer i|×|\layer{i-1}|}\), whose coefficients are computed as
\begin{equation*}
  S^x_{j,k} = （1 \mbox{~if~} \hat n_{i,j} = n_{i-1,k}, 0 \mbox{~otherwise}）
\end{equation*}
so that a 1 at row \(j\) and column \(k\) indicates that the output of neuron \(n_{i-1,k}\) as induced by \(x\) was selected by the layer as the value for \(\hat n_{i,j}\), \ie \(n_{i-1,k} = \max {s_j（n_{i-1,1}, …, n_{i-1,|\layer{i-1}|}）}\) as per Eq.~(\ref{eq:maxpooling-neuron-function}).
Note that multiple neurons of layer \layer {i-1} selected by a function \(s_j\) may be maxima, in which case there are more than one 1's in the corresponding row of \(S^x\).
Then, obtaining a new input that induces the same selection behaviour at layer \layer i boils down to enforcing either one of the following linear constraints:
\begin{equation*}
  （\hat n_{i,j} = n_{i-1,k}） \mbox{~if~} S^x_{j,k} = 1, （\hat n_{i,j} > n_{i-1,k}） \mbox{~otherwise.}
\end{equation*}

For an input neuron $n_{1,j}$, we have a variable $n_{1,j}$ and a constraint on its possible values based on the range of each input component (\eg pixel)
\begin{equation*}
n_{1,j}\in [a,b]\mbox.
\end{equation*}
We let \(\Constr[1,…,i]\N\) be the set of constraints that we obtain as described above for the neural network \N up to layer \layer i.

\subsubsection{Targeting Feature Intervals}
\label{sec:encod-targ-feat}

Further, a test target \Target{\subfeatspace i j k}, that aims at eliciting \LFI \(\subfeatspace i j k ∈ \Subfeatspaces i j\) in layer \layer i, directly translates into a pair of inequalities on neuron values for layer \layer i as follows:
\begin{equation}
  \label{eq:symbolic-feature-interval-target-constraint}
  \Constr{\subfeatspace i j k} ≝
  （\subfeatlb i j k ≤ λ_{i,j}（\hat n_{i,1}, …, \hat n_{i,|\layer i|}） < \subfeatub i j k）
\end{equation}
Observe that, if the feature mapping \(λ_{i,j}\) is linear, then the inequalities above are linear as well.
We straightforwardly extend the above construction to sets of target feature intervals for a layer \layer i: \Target{\DiscrFeatsElt i} translates into \Constr{\DiscrFeatsElt i}, that is the union (\ie the conjunction) of all inequalities built as per Eq.~(\ref{eq:symbolic-feature-interval-target-constraint}) for each interval in \DiscrFeatsElt i.

\subsubsection{Fulfilling BN-based Test Criteria}
\label{sec:fulfilling-bn-based-criteria}

When the goal of the symbolic analysis is to increase \BNFCov\N X coverage, the identification of a test target consists in finding an interval of \LF values that has a small probability of occurring according to \BNa\N X: this search boils down to identifying a target interval \subfeatspace i j k such that \(\MPr i {\subfeatspace i j k} < ε\), and building a target \(t = \Target{\subfeatspace i j k}\).

Finding a candidate test input that has the best chance of leading to an interesting solution for the LP problem is a hard problem.
Even more so when considering the potential losses in precision incurred by the underlying symbolic analyses.
We therefore make the simple assumption that an input \(x\) whose feature value \(λ_{i,j}∘\hat h_i(x)\) is close to any of the target interval boundaries is a ``good-enough bet''.
This assumption gives a simple heuristic for finding the candidate test input \(x\), which consists in searching for a test \(x ∈ X\), that minimises either \(|λ_{i,j}∘\hat\dimension_i(x) - \subfeatlb i j k|\) or \(|λ_{i,j}∘\hat\dimension_i(x) - \subfeatub i j k|\).

Then, the set of constraints \Constraints for specifying the LP problem for targeting \(t\) consists of:
\begin{equation}
  \Constraints = \Constr[1,…,i]\N ∪ \Constr{\subfeatspace i j k}
\end{equation}

\medskip

The case of \BNFdCov\N X is similar, as the test target identification consists in finding an entry \(\CPr i x {\subfeatspace i j k} {\DiscrFeatsElt{i-1}} < ε\) in any conditional probability table \(P_{\N,X}\BNode i j\).
One straightforwardly obtains \(t = \Target{｛\subfeatspace i j k｝∪\DiscrFeatsElt{i-1}}\), that aims at covering both the interval \subfeatspace i j k and a condition on \LFs of layer \layer{i-1} for which the probability of finding an input in \(X\) that exhibits \subfeatspace i j k is very small.
The selection of a test input can also be done according to a heuristic search like the one above, and the resulting set of constraints is:
\begin{equation}
  \Constraints = \Constr[1,…,i]\N ∪
  \Constr{\DiscrFeatsElt{i-1}} ∪ \Constr{\subfeatspace i j k}
\end{equation}

\subsection{Helping Convergence}
\label{sec:helping-convergence}

In the description of the algorithm above, we have hinted at a practical measure that we use to counter a consequence of the loss of precision that is induced by dimensionality reduction.
Indeed, although the latter comes in addition to the imprecision that is notably due to linearly encoding non-linear behaviours such as max-pooling layers---discussed in Section~\ref{sec:convergence}---, our preliminary empirical evaluations have shown us that the aforementioned imprecision has an adverse effect on the ability of our adapted algorithm to fulfil our BN-based criteria.

For instance, given a test target \(t = \Target{\subfeatspace i j k}\) and a candidate test input \(x\), the solution \(x'\) of the resulting LP program often leads to a \LF valuation \(λ_{i,j}∘\hat h_i(x')\) that does not belong to \subfeatspace i j k: in other words, the actual behaviour at \layer i subject to input \(x'\) does not satisfy Eq.~(\ref{eq:symbolic-feature-interval-target-constraint}) due to the approximations induced by the feature mapping \(λ_{i,j}\).
Even worse, \(x'\) may actually be ``further'' on the \featcomp i j line from the target than \(x\): \ie \(|t - \hat h_i(x)| < |t - \hat h_i(x)|\), where \(|t - y| ≝ \min（|λ_{i,j}(y) - \subfeatlb i j k|, |λ_{i,j}(y) - \subfeatub i j k|）\) measures the distance between \(t\) and the \LF value induced by neuron values \(y\) for layer \layer i.

On a purely theoretical level, we acknowledge that this convergence issue is an intrinsic limitation of our technique, as it may be possible that none of the constructed LP problems ever have any solution that improves the desired coverage metric.
Still, we can identify ways to counter this issue in practice.
First, we can observe that rejecting any newly generated input on the grounds that it does not fulfil the test target quickly leads to the inability to find new candidate test inputs for any unmet test target.
Instead, considering such inputs as legitimate new test cases even though they do not improve coverage helps populating the set of potential candidate tests inputs: they indeed help further explorations of the input space.
Further, keeping only a new input \(x'\) if it improves over its original test \(x\) \wrt the target \(t\) may naturally lead to the construction of a set of test cases that is ``closer'' to satisfying the criterion (although this is not a requirement for actually helping convergence).
We validate these claims further with dedicated experiments in Section~\ref{sec:actual-executions}.

At last, we posit that further constraining the set of admissible solutions of the LP programs indirectly gears the set of generated inputs towards cases that meet the desired targets.
In particular, we assume that constraining non-targeted \LFs so that they replicate the respective values induced by \(x\) reduces the over-approximations, \ie increases the precision of the symbolic analysis.
We achieve this by introducing the additional set of linear constraints
\begin{equation}
  \label{eq:eq:symbolic-other-components-constraint}
  \Replic x {\subfeatspace i j k} ≝ ⋃_{j' ∈ ｛1,…,\lowdim i｝, j' ≠ j}
  （λ_{i,j'}（\hat n_{i,1}, …, \hat n_{i,|\layer i|}） = λ_{i,j'}∘\hat h_i(x)）
\end{equation}
into the set \Constraints built for targeting \Target{\subfeatspace i j k} from an input \(x\) (in the case of a BN-based feature criterion).
We empirically support this assumption using a dedicated set of experiments in the next Section.

\section{Implementation and Evaluations}
\label{sec:implem-n-evaluations}

We have augmented the \DeepConcolic tool\footnote{Available at \url{https://github.com/TrustAI/DeepConcolic/}.} in order to experimentally validate the practicality and efficiency of our new coverage metrics, as well as our adaptation of concolic test case generation.
With the following experiments, we want to assess whether our BN-based approach meets the needs to define semantics-based coverage criteria, that go beyond the low-level structure of neural networks.

We have implemented multiple strategies for linear dimensionality reduction and discretisation of each feature component.
We have notably included PCA (with or without pre-scaling of activation values) and ICA (\cf Section~\ref{sec:dimens-reduct-via}), as well as several strategies for computing discrete partitions (\cf Section~\ref{sec:discr-strat}).

The wide range of combinations that all these strategies offer, even more when a distinct choice can be made regarding each individual layer, makes the complete study of this choice a challenging task.
As in the first step we mainly want to assess the practicality of our BN-based approach for defining coverage criteria, for now we will report on some results for broad families of combinations of strategies.
All experiments were carried out on the DNNs \Nms and \Nmx described in \tablename s~\ref{tab:dnn-Nms} and~\ref{tab:dnn-Nmx}, that target the classification of the handwritten digits from the MNIST dataset.

\begin{leaveout}
  \itodo{The experimental results reported in the subsection below (in red) are still too inconclusive alas…}

\newcommand\Nmnist{\ensuremath{\N_{\mathsf{MNIST}}}\xspace}%

\begin{redenv}
\subsection{Evaluating Dimensionality Reduction and Discretisation
  Strategies}

We have re-used a 
15-layer DNN provided with \DeepConcolic to conduct an evaluation for
dimensionality reduction and discretisation strategies.
This DNN features multiple convolutional, max-pooling, and dense
layers, to achieve the classification of handwritten digits given as
black-and-white images of 28×28 pixels.
We refer to this DNN as \Nmnist in the following.
\newcommand\Xfull{\ensuremath{X_{\mathit{all}}}\xspace}%
\newcommand\Xnobar{\ensuremath{X_{\mathit{nobar}}}\xspace}%
We used a sample of 1000 inputs known to be correctly classified by
\Nmnist to construct \Xtrain, and then derived the structure of the
Bayesian Networks (\ie feature components and associated intervals,
which translate into nodes and uniform probability tables in BN terms)
by means of the strategies listed in Section~\ref{sec:discr-strat}:
binarisation, \(k\)-bins-uniform, \(k\)-bins-quantile,
\(k\)-bins-uniform-extended, and \(k\)-bins-quantile-extended, (with
\(k\) ranging in \(｛1,…,5｝\)).

To check whether the coverage metrics given in
Def.~\ref{def:bn-based-feature-coverage} and
Def.~\ref{def:bn-based-feature-dependence-coverage} are effective, we
have computed the coverage obtained \wrt each respective BN on two
test sets:
\begin{description}
\item[\Xfull] a sample of 20 test inputs that features every digit;
\item[\Xnobar] a sample of 20 test inputs that does \emph{not} feature
  any digit that is usually represented with a straight vertical bar
  (\ie `1', `4', and `7').
\end{description}
The idea underlying our choice for comparing the coverage we obtain
for these two test sets is that the classification of \Xnobar should
involve less semantic assumptions, or feature relationships, than the
classification of \Xfull.
%
This way, we can compare the coverage obtained by each combination of
strategies for \Xfull against \Xnobar, and assess whether the coverage
obtained for the former is indeed larger than that obtained for the
latter.
To ease comparisons, we will use square scatter plots to represent the
results, where the horizontal (resp. vertical) axis marks the coverage
obtained for \Xnobar (resp. \Xfull).
Therefore, in all the plots of this Section, the dots that lie above
the dashed grey diagonal line correspond to a combination of
strategies that leads to the expected result: the coverage obtained
for the test set \Xnobar is lower than the coverage obtained for
\Xfull.

Note that we rely on a single coverage result for each pair of
strategy and input test set: this simplification is motivated by the
consistent results we obtained when running the same coverage
computations several times, and the need for presenting our comparison
results in a concise way.

We first report the comparison plots that we obtain for each pair of
coverage metric and test inputs in \figurename
s~\ref{fig:mnist-0-2-5-7-11-13-15} and~\ref{fig:mnist-11-13-15}.
In each subplot (a) and (b) of these two Figures, the size of the
markers represents the number of features that have been extracted for
each layer (which ranges from 1 to 4); these plots do not represent
the number of individual intervals per component.
Note that many markers are stacked onto one another on the diagonal
line in the (a) plots, about \(\BNFCov{\Nmnist}{⋅}\).
Still, these plots indicate that only a few strategies match our
requirements for distinguishing untested features.
However, we observe from the two plots (b) that are related to
feature-dependence coverage in both Figure, that focusing on
middle-to-deepest layers, and using the PCA technique to extract a
single feature per layer, tends to lead to the expected coverage
results for this metric.

%
%
%
We complete the latter result by representing in the plots (c) and (d)
of \figurename s~\ref{fig:mnist-0-2-5-7-11-13-15}
and~\ref{fig:mnist-11-13-15} the respective number of extracted
feature component per considered layer, as well as the number of
interval per extracted feature components.
We observe that there is an upper limit to the number of feature to
extract (such as 5 in this case) above which high coverage values are
always obtained, even for \Xnobar: in general the coverage metric for
the same inputs increases with the number of extracted components for
each layer.
Furthermore, the number of interval per component does not appear to
be a relevant factor for discriminating semantics assumptions that are
not covered by a set of test inputs.


\end{redenv}
\end{leaveout}

\subsection{Concolic Testing for BN-based Criteria}
\label{sec:actual-executions}

We have run the extended concolic testing tool on \Nms and \Nmx by focusing on their dense layers (named \texttt{dense} and \texttt{dense\_1} in both cases), with varying numbers of extracted feature components for each one of these two layers, using various algorithms for feature extraction, and/or changing the discretisation strategies.
Each run comprised at most 100 iterations, and was initialised with uniformly drawn test sets \(X_0\) of 10, 100, or 1000 correctly classified inputs.
\begin{figure}%
  \centering%
  \newlength\refw%
  \setlength\refw{28px}%
  \newcounter{y}%
  \setcounter{y}{0}%
  \begin{tikzpicture}%
    \node[inner sep=0pt, anchor=north] (a) {
      \includegraphics[width=3\refw]{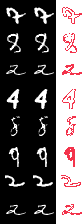}%
    };
    \foreach \lbl in {
      4,
      2,
      1,
      9,
      5,
      7,
      4,
      1
    }{
      \node[yshift=-\arabic{y}\refw, color = red, anchor=north east,
            inner sep = 1pt] at ($(a.north west)+(2\refw, 0)$) {\lbl};
      \stepcounter{y}%
    }
  \end{tikzpicture}
  \caption{Adversarial examples found by achieving a BN-based feature coverage criterion for all dense layers of \Nms.
    Each row includes, from left to right, the original test image, the adversarial example (with its new classification label in overlay), and an image that emphasises every pixel that differs.}
  \label{fig:Nms-bfc-adversarials}%
\end{figure}

\begin{figure}%
  \centering%
  \def\heigh{3cm}%
  \includegraphics[height=\heigh]{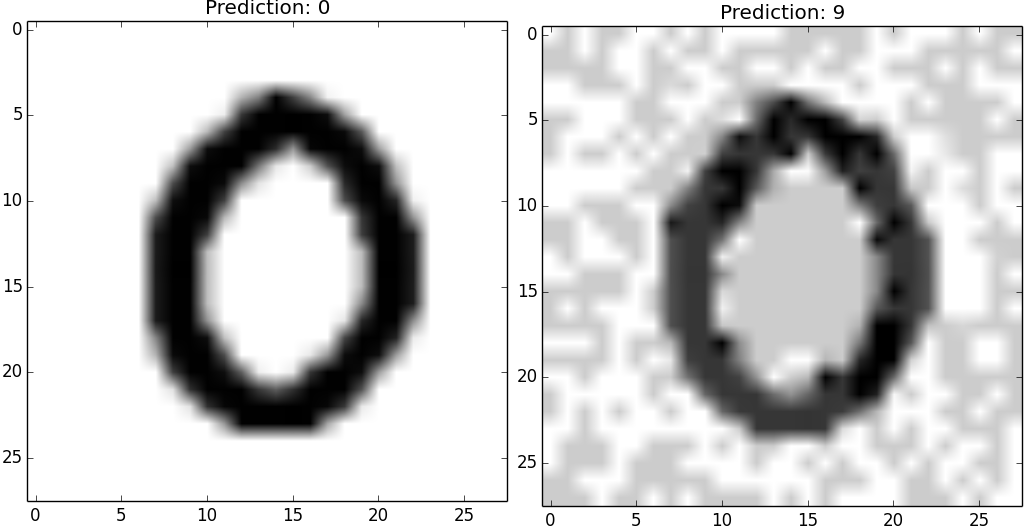}%
  \caption{Adversarial example found by \DeepConcolic while seeking a structural coverage criterion; this is borrowed from \citet{sun2018testing}.}%
  \label{fig:ns-adversarials-structural}%
\end{figure}%
We give in \figurename~\ref{fig:Nms-bfc-adversarials} some adversarial examples that were found during the experiments on \Nms and when targeting the BN-based feature coverage criterion.
For comparison purposes, we also show in \figurename~\ref{fig:ns-adversarials-structural} an adversarial example found by \DeepConcolic to achieve a structural criterion (\ie expressed on individual neuron activations instead of on \LFs).
We observe that, even though the distance metric used in all three cases was the same (\(L_∞\)), the example found through structural coverage appears to involve small-grain perturbations of pixels.
On the contrary, our adversarial examples, and in general every generated test, exhibit differences that seem to precisely indicate where the input images need to be perturbed to alter the induced value for the respective targeted \LF.

We note that most adversarial examples were found during the first 30 to 40 iterations of the test case generation algorithm.
Further, most adversarial examples were found when using \textit{extended} discretisation strategies (\cf Section~\ref{sec:discr-strat}), where two additional left- and right- open intervals that do not contain any value that is induced by the training dataset are used for partitioning \LF components.
This suggests that attempting to identify intervals in a \LF component \featcomp i j that do not exhibit any feature value (\ie finding intervals whose intersection with the set \(λ_{i,j}∘\hat\dimension_i(\Xtrain)\) is empty, where \Xtrain is the training dataset), seems a suitable strategy to find corner-cases for DNNs.

\begin{figure}
  \centering
  \input{fig/mnist_small-bfc-summary-per-X0.pgf}
  \caption{Representation of 360 traces of up to 100 iterations of test case generation by \DeepConcolic targeting BN-based feature coverage, for various sizes of initial test sets \(|X_0| ∈ ｛10,100,1000｝\).
    Red and blue lines respectively indicate runs with ICA and PCA-based feature extractions.
    Each box in the bottom row shows the overall distribution of initial \(\BNFCov\Nms{X_0}\) and the respective final coverage \(\BNFCov\Nms{X_{100}}\).
  }
  \label{fig:Nms-bfc-summary}
\end{figure}
\begin{figure}
  \centering
  \input{fig/mnist_small-bfdc-summary-per-X0.pgf}
  \caption{Representation of 360 traces of up to 100 iterations of
    test case generation by \DeepConcolic targeting BN-based
    feature-dependence coverage, for various sizes of initial test
    sets \(|X_0| ∈ ｛10,100,1000｝\).
    Red and blue lines respectively indicate runs with ICA and
    PCA-based feature extractions.
    Each box in the bottom row shows the overall distribution of
    initial \protect\(\BNFxCov\Nms{X_0}\) and the respective final coverage
    \protect\(\BNFxCov\Nms{X_{100}}\).
  }
  \label{fig:Nms-bfxc-summary}
\end{figure}
Let us now turn to a discussion on the overall behaviours of the concolic testing approach delineated in Section~\ref{sec:cover-guid-test} in regard to the BN-based coverage metrics defined in Section~\ref{sec:bn-based-coverage}.
In particular, we want to assess the impact of the size of the initial test set (\(X_0 \)) on the ability of \DeepConcolic to generate new sets of inputs that achieve high coverage.

We summarise the runs for BN-based feature coverage in \figurename~\ref{fig:Nms-bfc-summary}.
We first observe that, overall, between 10\% to 60\% of iterations produce new test inputs, and that the increase in the size of \(X_i\) is not correlated with the corresponding feature coverage.
Instead, the coverage tends to increase in discrete steps, and often reaches plateaus for significant numbers of iterations.
The former observation is a direct consequence of the discrete nature of the structure---the BN---based on which we defined the coverage metrics.
This behaviour also gives a great illustration of our concerns for convergence raised in Section~\ref{sec:helping-convergence}.
We also note that neither PCA (in blue) or ICA (in red) seem to have any impact on the number of generated tests or the achieved coverage.

Further, the bottom row in \figurename~\ref{fig:Nms-bfc-summary} indicates that 100 iterations already leads to increases in achieved coverage when the initial size of the test set \(X_0\) is sufficiently large.
This confirms that the size of the test set helps our heuristics in identifying good candidate test inputs for a given test target.

We now turn to our results for BN-based feature-dependence coverage, summarised in \figurename~\ref{fig:Nms-bfxc-summary}; note that the coverage measure that is plotted is the one defined in Eq.~(\ref{eq:bfxc}), which is the combined version of pure feature coverage and feature-dependence.
We first observe that many runs terminate before reaching 100 iterations.
This can be tracked down to two root causes:
\begin{enumerate*}[(i)]
\item the feature-dependence coverage reaches 100\% (while full feature coverage is not achieved), which basically means that every conditional probability table entry is greater than ε whereas some entries of marginal probability tables are still negligible;
\item no new pair (test target, candidate test input) can be identified (\ie every such pair has already been attempted).
\end{enumerate*}
Still, we can note an overall increase in coverage for \(|X_0| = 10\).
This suggests that a dedicated heuristics is required to achieve both feature and feature-dependence coverage in combination.

\subsection{Convergence \& Precision Aspects}
\label{sec:convergence-aspects}

In Section~\ref{sec:helping-convergence}, we posited that constraining the LP program in such as way that \LF components that are not related to the target replicate the values induced by the source input \(x\), alleviates the convergence issues.
We have again used \Nms to assess this assumption.
We have conducted two sets of experiments, that each consisted in a set of 40 runs with each parameter uniformly drawn from a set of pre-defined values for the number of extracted features per layer, discretisation strategy, number of bins, etc.
Each set of runs differs in that for a single one of them the counter measure detailed in Eq.~(\ref{eq:eq:symbolic-other-components-constraint}) was employed.

\begin{figure}
  \centering
  \input{fig/mnist_small-bfc-nofixed-ica-dist-n-progress.pgf}\\
  (a) without replication of values for non-targeted \LF components\par
  \input{fig/mnist_small-bfc-cfixed-ica-dist-n-progress.pgf}\\
  (b) with replication of values for non-targeted \LF components\par
  \caption{Assessing the impact of constraining non-targeted \LF components on progress, using \Nms.
    On the left: histograms representing the distribution of all distances to the closest bounds of target \LFIs during 40 runs of up to 100 iterations (with extended discretisation strategy and bins uniformly drawn from a fixed set of values) of test case generation by \DeepConcolic targeting BN-based feature coverage when ICA was used as feature extraction technique---PCA gives similar-looking results, although it does not map to feature components that are comparable with each other.
    On the right: similar histograms showing the difference between the distance of a source test input \(x\) and a newly generated one \(x'\): \(\delta > 0\) indicates that the generated input is closer to the target \LFI than its source test.
    Contrary to the top row (a), the bottom row (b) shows distributions for \(d\) and δ obtained when the counter measure of Eq.~(\ref{eq:eq:symbolic-other-components-constraint}) was enabled.}
  \label{fig:Nms-bfc--ica-feature-dist-n-progress}
\end{figure}
We plot in \figurename~\ref{fig:Nms-bfc--ica-feature-dist-n-progress} the resulting distributions of: distance from generated inputs to target (\(d\) --- on the respective components line), along with the similar histograms showing the difference between the distance of a source test input \(x\) and a newly generated one \(x'\): \(\delta > 0\) indicates that the generated input is closer to the target \LFI than its source test.
We can mainly observe that using the counter measure has no noticeable impact on our measure of progress (δ), yet it seems to gear generated inputs closer to their respective targets overall.
We made similar observations when PCA was used for feature extraction, and when performing the same experiments with other DNNs.

\begin{figure}
  \centering
  \input{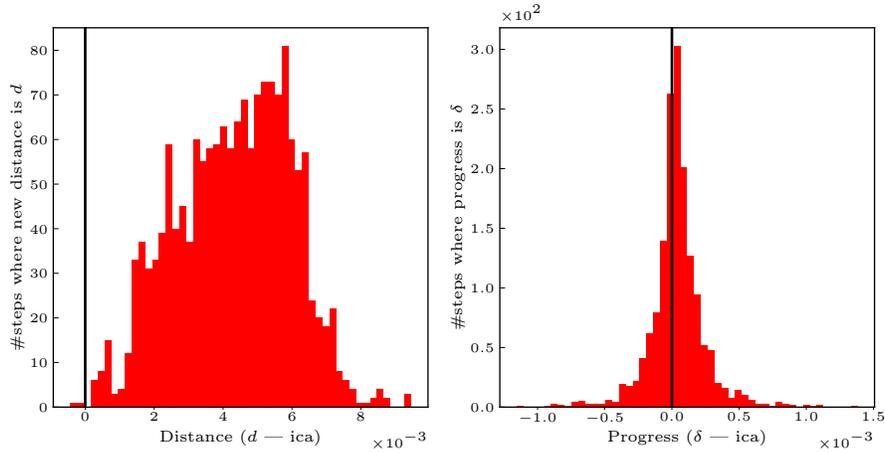}\\
  (a) without replication of values for non-targeted \LF
  components\par
  \input{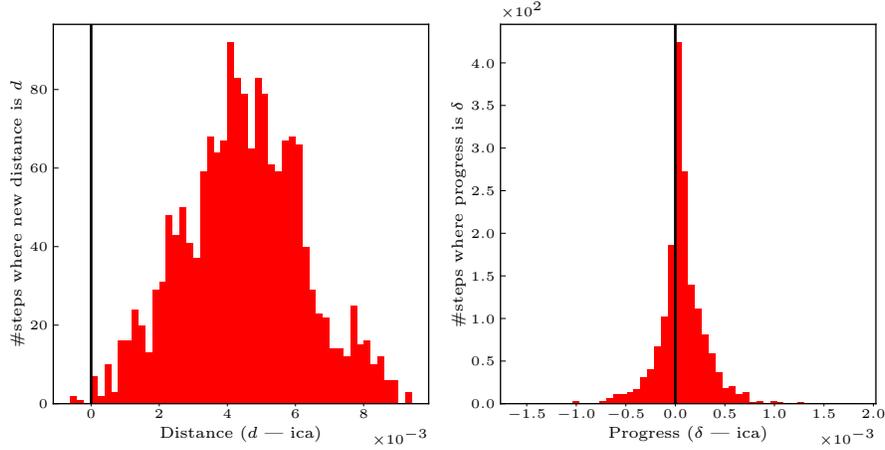}\\
  (b) with replication of values for no-targeted \LF components\par
  \caption{Assessing the impact of max-pooling layers, and constraining non-targeted \LF components on progress, using \Nmx.
    See \figurename~\ref{fig:Nms-bfc--ica-feature-dist-n-progress} for a description of the histograms.}
  \label{fig:Nmx-bfc--ica-feature-dist-n-progress}
\end{figure}
We could note, however, that the above impact seems to be barely noticeable in the case of networks with max-pooling layers.
We illustrate this by showing in \figurename~\ref{fig:Nmx-bfc--ica-feature-dist-n-progress} the histograms obtained from an identical set of experiments for \Nmx.
This suggests that the benefits of the counter measure is attenuated by the loss in precision that is incurred by our symbolic encoding of such intrinsically non-linear layers.

\section{Related Work}\label{sec:related}
The investigation of techniques to analyse the weaknesses of deep learning has been a very popular research domain in the past few years.
In the following, we briefly review a few directions that are closely related to this paper.
Please refer to~\cite{HUANG2020100270} for a comprehensive review.

\paragraph{Testing} includes methods to generate a set of test cases and use the generated test cases to evaluate the reliability (or other properties) of deep learning. There are a number of ways to determine how the test cases are generated, including \eg fuzzing~\cite{10.1145/3236024.3264835}, coverage metrics~\cite{SHKSHA2019,huang2020coverage}, symbolic execution~\cite{DBLP:journals/corr/abs-1807-10439,DBLP:journals/corr/abs-1803-04792}, concolic testing~\cite{sun2018concolic}, etc.

\paragraph{Verification} includes algorithms to   check if a deep learning model satisfies a property. Existing verification algorithms can be roughly categorised into constraint-solving based methods~%
\cite{katz2017reluplex}, abstract interpretation based methods~%
\cite{gehr2018ai,10.1007/978-3-030-32304-2_15,Singh2019DeepPoly}, global optimisation based methods~%
\cite{HKWW2017,RHK2018,ijcai2019-824}, and game-based methods~\cite{10.1007/978-3-319-89960-2_22,wu2018game}.

\paragraph{Abstraction}
There are a few existing methods on abstraction of deep learning.
For example, in \cite{DBLP:journals/corr/abs-1809-06573}, a Boolean abstraction on the ReLU activation pattern of some specific layer is considered and monitored.
Conversely of Boolean abstraction, \citet{DBLP:journals/corr/abs-1911-09032} and \citet{cheng2020provablyrobust} consider box abstractions.
The qualitative box abstraction is enhanced by \citet{lukina2020unknown} to identify unexpected inputs with a quantitative abstraction of box distance, which measures the minimum distance to one of the class centres.

\paragraph{Probabilistic Assessment} unlike testing and verification which are mainly to determine the existence of ``bugs'' (\ie counterexamples to the satisfiability of desirable properties) in the deep learning model, assesses the satisfiability of a property in a probabilistic way, by \eg summarising over sampling results.
Due to the unknown ground truth over the underlying distribution, this approach usually requires to either make an assumption over the underlying distribution~\cite{weng2018evaluating} or learn the distribution~\cite{zhao_safety_2021}.
Note that the obtained probabilistic assessment can be bounded~%
\cite{bishop_conservative_2014,jin2020does}.

\paragraph{Certification and Safety Arguments}
While the above techniques may compute the evidence to the (un)satisfiability of a property, the properties they work with are mainly low-level ones, such as the point-wise robustness which only regards robustness \wrt a given input.
More investigations are needed to understand if and how the evidence we can gain against these low-level specifications, can contribute to the claim of higher-level specifications such as ``a deep learning model can be free from failure for the next $k$ inferences''~\cite{zhao_safety_2020}.
One step further, it might be interesting to understand the safety issues in learning-enabled autonomous systems~\cite{9196932,9340720}.

\section{Conclusions}
\label{sec:conclusions}

In this paper, we have suggested a new way to abstract the successive layers of a trained DNN by means of a Bayesian Network.
The abstraction relation between these two objects is parameterised by a linear feature extraction mapping and a discretisation strategy for the features.

The constructed model can also be used to assess the quality of a set of test inputs \wrt both the amount of individual features it covers, and assumed relationships between features.
We have exercised this ability by extending the test case generation tool \DeepConcolic with our new BN-based coverage criteria.
We have carried out extensive experiments to assess the effectiveness of this tool in generating new inputs despite the approximations that stem from the feature extraction techniques, as well as the linear encoding of non-linear functional behaviours (like max-pooling layers).
Our experimental results show that the new tool is able to efficiently generate new test inputs, including adversarial examples, and generates new inputs by emphasising portions of learnt features to legitimate inputs.
However, the formulation of our coverage metrics suffers from the discrete structure of the underlying BN, and fails to precisely account for any small-scale improvement towards fulfilling the associated criteria.

\bibliography{references}
\bibliographystyle{plainnat}

\end{document}

%% file: macro.tex
\newcommand{\commentout}[1]{}

\newcommand{\network}{{\cal N}\xspace}

\DeclareMathOperator*{\argmax}{arg\,max}

\usepackage{xifthen}
\newcommand\lambdaone[1]{\ensuremath{\ifthenelse{\isempty{#1}}{}{（#1）}}}
\newcommand\lambdatwo[2]{\ensuremath{\ifthenelse{\isempty{#1#2}}{}{（#1, #2）}}}

\newcommand{\inputdomain}{\ensuremath{\DD_X}\xspace}
\newcommand{\outputdomain}{\ensuremath{\DD_Y}\xspace}
\newcommand\layerdom[1]{\ensuremath{\LL_{
      {#1}}}\xspace}
\newcommand\featdom[1]{\ensuremath{\FF_{
      {#1}}}\xspace}
\newcommand\featcomp[2]{\ensuremath{\FF_{
      {#1},{#2}}}\xspace}

\newcommand{\itodo}[2][]{\todo[inline,caption={2do}, #1]{%
    \begin{minipage}{\textwidth-4pt}#2\end{minipage}}}

\newenvironment{redenv}{\color{red}}{}

\newcommand{\adhoc}		{{\itshape	 ad hoc\/}\xspace}

\newcommand{\cf}		{{\itshape	      cf.}\xspace}

\newcommand{\eg}		{{\itshape	    e.g.,}\xspace}

\newcommand{\ie}		{{\itshape	    i.e.,}\xspace}
\newcommand{\perse}		{{\itshape	   per se}\xspace}
\newcommand{\wrt}		{{\itshape	   w.r.t.}\xspace}

\newcommand{\DeepConcolic}{\textrm{DeepConcolic}\xspace}

\renewcommand\P{\ensuremath{\mathcal P}\xspace}%
\newcommand\B{\ensuremath{\mathcal B}\xspace}%
\newcommand\N{\ensuremath{\mathcal N}\xspace}%
\newcommand\DD{\ensuremath{\mathbb D}\xspace}%
\newcommand\FF{\ensuremath{\mathbb F}\xspace}%
\newcommand\LL{\ensuremath{\mathbb L}\xspace}%
\newcommand\RR{\ensuremath{\mathbb R}\xspace}%

\newcommand\Layers{\ensuremath{\mathsf{Layers}}\xspace}
\newcommand\layer[1]{\ensuremath{l_{#1}}\xspace}

\newcommand\Features[1]{\ensuremath{Λ
    _{#1}}\xspace}

\newcommand{\dist}{{\cal D}}

\newcommand{\dimension}{h}
\newcommand\lowdim[1]{\ensuremath{|Λ_{#1}|}\xspace}

\newcommand\DiscrFeatsSpace[1]{\ensuremath{\FF^{♯}_{#1}}\xspace}
\newcommand\DiscrFeatsElt[1]{\ensuremath{F^{♯}_{#1}}\xspace}
\newcommand\AllSubFeatSpaces[1]{\ensuremath{\FF^{♯}_{#1}}\xspace}

\newcommand\Subfeatspaces[2]{\ensuremath{\FF^{♯}_{#1,#2}}\xspace}
\newcommand\subfeatspace[3]{\ensuremath{f^{♯#3}_{#1,#2}}\xspace}
\newcommand\subfeatlb[3]{\ensuremath{f^{[#3}_{#1,#2}}\xspace}
\newcommand\subfeatub[3]{\ensuremath{f^{#3[}_{#1,#2}}\xspace}
\newcommand\featsubsp[2]{\ensuremath{\mathsf{Discr}^{♯}_{#1,#2}}\xspace}

\renewcommand\P[1]{\ensuremath{\mathrm{Pr}\!\left(#1\right)}\xspace}

\newcommand\LF[1][h]{#1idden feature\xspace}
\newcommand\LFs[1][h]{#1idden features\xspace}
\newcommand\LFI[1][h]{#1idden feature interval\xspace}
\newcommand\LFIs[1][h]{#1idden feature intervals\xspace}

\newcommand\Nms{\ensuremath{\N_{\mathsf{ms}}}\xspace}%

\newcommand\Nmx{\ensuremath{\N_{\mathsf{mx}}}\xspace}%

\newcommand\Target[1]{\ensuremath{\mathit{Target}（#1）}\xspace}%
\newcommand\Constr[2][]{\ensuremath{\mathit{Constr}_{#1}（#2）}\xspace}%
\newcommand\Replic[2]{\ensuremath{\mathit{Replic}_{#1}（#2）}\xspace}%
\newcommand\LInf[2]{\ensuremath{\|#1 - #2\|_{∞}}\xspace}%

\newcommand\Xtrain{\ensuremath{X_{\mathit{train}}}\xspace}%


%% file: mylists.tex
\setlist{noitemsep}
\setlist[1]{labelindent=\parindent} 
\newlist{compactitem}{itemize}{4}
\setlist[compactitem,1]{nolistsep,label=\textbullet}
\setlist[description]{font=\mdseries\itshape}
\newlist{mathdesc}{description}{4}
\newlist{mathpardesc}{description}{4}
\newlist{mathpardesc*}{description}{4}
\newlist{mathpardesc**}{description*}{4}
\newcommand*{\keymathbox}[1]{%
  \mdseries%
  \upshape%
  \setlength{\fboxsep}{.4pt}%
  \fcolorbox{gray}{white}{%
    \strut\ensuremath{#1}%
  }%
}%
\setlist[mathdesc]{format=\ensuremath}
\setlist[mathpardesc]{format=\keymathbox,
  leftmargin=\parindent,
  labelindent=0pt,
}
\setlist[mathpardesc*]{format=\keymathbox,
  nosep,
  leftmargin=0pt,
  labelindent=\parindent,
}
\setlist[mathpardesc**]{format=\keymathbox,mode=unboxed}

\newlist{bolddescr}{description}{4}
\newcommand*{\mybolddescritem}[1]{\bfseries\upshape{#1}:}%
\setlist[bolddescr]{style=sameline, nosep, format=\mybolddescritem,
  leftmargin=0pt, labelindent=0pt,
}

\newcommand*\maindescfmt[1]{\upshape\bfseries #1:}%
\newlist{maindesc}{description}{4}
\setlist[maindesc]{%
  nosep,
  format=\maindescfmt,
  style=nextline,
  labelindent=0pt,
  leftmargin=\parindent,
}

\newlist{subdesc}{description}{4}
\setlist[subdesc]{%
  nosep,
  style=sameline,
  format=\mdseries\itshape,
  leftmargin=\parindent,
}

%% file: tikzconf.tex
\usetikzlibrary{fit,calc}
\usetikzlibrary{arrows, arrows.meta}
\usetikzlibrary{decorations, decorations.pathmorphing}
\usetikzlibrary{intersections}
\tikzset{
  every path/.style = {
    line width=1pt,
    cap=round,
    join=round,
  },
  rho-highlight/.style = {
    rectangle,
    rounded corners=2pt,
    draw,
    thin,
    inner sep=0pt,
    outer sep=0pt,
  },
  rho-annot-arrow/.style = {
    thin,
    -stealth,
  },
}

%% file: diagrams/feat-space-discr.tikz
\begingroup
\smaller
\begin{tikzpicture}
  \colorlet{featcol1}{blue!66!black}
  \colorlet{featcol2}{red!66!black}

  \colorlet{rangecol}{black!40!white}
  \colorlet{rangecoltxt}{black!30!rangecol}
  \tikzstyle{full-range} = [
  shorten >=-2pt]
  \tikzstyle{li1-range} = [{Arc Barb[round,slant=-0.2,scale=.8]}-{Arc Barb[reversed,slant=-0.2,scale=.8]},full-range,draw = featcol1!50!white]
  \tikzstyle{li2-range} = [{Arc Barb[round,slant=0.33,scale=.8]}-{Arc Barb[reversed,slant=0.33,scale=.8]},full-range, draw = featcol2!50!white]
  \tikzstyle{li2-range'} = [{Arc Barb[round,slant=0.33,scale=.8]}-,draw = featcol2!50!white]

  \tikzstyle{featsubspace1} = [outer sep = 0pt, inner sep = 1pt, anchor = south, xshift=1mm, featcol1]
  \tikzstyle{featsubspace2} = [outer sep = 0pt, inner sep = 1pt, anchor = south east, xshift=2mm, featcol2]

  \coordinate (li1-beg) at (-8:-3.4);
  \coordinate (li1-1) at (-8:-2.6);
  \coordinate (li1-2) at (-8:1.5);
  \coordinate (li1-3) at (-8:4);
  \coordinate (li1-end) at (-8:5.2);

  \coordinate (li2-beg) at (20:-3.2);
  \coordinate (li2-1) at (20:-1.5);
  \coordinate (li2-2) at (20:2.1);
  \coordinate (li2-end) at (20:4.5);


  \draw[->,name path=li1] (li1-beg) -- (li1-end) node[anchor = west,
  inner sep = 1pt] {\larger\featcomp i 1};

  \foreach \j in {1,...,3} {
    \draw[dotted, draw = rangecol] ([yshift=1mm]li1-\j) -- ++(0,-5mm) node[anchor = north,inner sep=.1pt,rangecoltxt] {%
      \pgfmathtruncatemacro\jj{\j+1}
      \smaller\smaller\smaller\subfeatlb i 1 {\jj}%
    };
    \fill[featcol1] (li1-\j) circle (1.6pt);
  }
  \draw[li1-range] ([yshift=-2mm]li1-1) -- ([yshift=-2mm]li1-2)
  ;


  \draw[->,name path=li2] (li2-beg) -- (li2-end) node[anchor = west,
  inner sep = 1pt] {\larger\featcomp i 2};

  \fill[name intersections={of=li2 and li1, by=x}] (x) circle (1.2pt);
  \foreach \j in {1,...,2} {
    \draw[dotted, draw = rangecol] ([yshift=1mm]li2-\j) -- ++(0,-5mm) node[anchor = north,inner sep=.1pt,rangecoltxt] {%
      \pgfmathtruncatemacro\jj{\j+1}
      \smaller\smaller\smaller\subfeatlb i 2 {\jj}%
    };
    \fill[featcol2] (li2-\j) circle (1.6pt);
  }

  \coordinate (x) at ($(li2-2)!2/3!(li2-end)$);
  \coordinate (y) at ($(li2-2)!5/6!(li2-end)$);
  \draw[li2-range'] ([yshift=-2mm]li2-2) -- ([yshift=-2mm]x);
  \draw ([yshift=-2mm]x) [densely dashed, draw = featcol2!50!white] -- ([yshift=-2mm]y);
  \draw ([yshift=-2mm]y) [dotted, draw = featcol2!50!white] -- ([yshift=-2mm]li2-end);


  \node[featsubspace1] at ($(li1-beg)!.5!(li1-1)$) {\subfeatspace i 1 1};
  \node[featsubspace1] at ($(li1-1)!.42!(li1-2)$) {\subfeatspace i 1 2};
  \node[featsubspace1] at ($(li1-2)!.52!(li1-3)$) {\subfeatspace i 1 3};
  \node[featsubspace1] at ($(li1-3)!.42!(li1-end)$) {\subfeatspace i 1 4};

  \node[featsubspace2] at ($(li2-beg)!.3!(li2-1)$) {\subfeatspace i 2 1};
  \node[featsubspace2] at ($(li2-1)!.6!(li2-2)$) {\subfeatspace i 2 2};
  \node[featsubspace2] at ($(li2-2)!.52!(li2-end)$) {\subfeatspace i 2 3};


  \colorlet{actcol}{green!66!black}
  \colorlet{actspacecol}{actcol!20!white}

  \coordinate (fx1) at (-8:-1.4);
  \coordinate (fx2) at (20:2.5);
  \coordinate (fx) at ($(fx1)+(fx2)$);
  \coordinate (hx) at (85:2.8);
  \draw[fill = actspacecol, draw = none] (hx)
  ellipse [xshift = -8mm, yshift = -2.4mm, x radius = 1.6cm, y radius = 8mm, rotate=10];
  \node[anchor = north east, xshift=-1cm, yshift=-1mm,
  black!40!actcol] at (hx) {\larger\layerdom i};
  \fill[black] (fx1) circle (2pt);
  \fill[black] (fx2) circle (2pt);
  \node [anchor = south] at (hx) {\(\hat h_i(x)\)};
  \draw[dotted, featcol1, {Latex}-] (fx1) to[out = 90, in = -130, edge node = {
    node [sloped, above, featcol1, pos = .3] {\(λ_{i,1}
      \)}}] (hx);
  \draw[dotted, featcol2, {Latex}-] (fx2) to[out = 90, in = -30, edge node = {
    node [sloped, above, featcol2] {\(λ_{i,2}
      \)}}] (hx);
  \fill[black] (hx) circle (2pt);
\end{tikzpicture}
\endgroup
 

%% file: main.bbl
\begin{thebibliography}{41}
\providecommand{\natexlab}[1]{#1}
\providecommand{\url}[1]{\texttt{#1}}
\expandafter\ifx\csname urlstyle\endcsname\relax
  \providecommand{\doi}[1]{doi: #1}\else
  \providecommand{\doi}{doi: \begingroup \urlstyle{rm}\Url}\fi

\bibitem[Ashmore and Hill(2018)]{box-clever}
Rob Ashmore and Matthew Hill.
\newblock Boxing clever: Practical techniques for gaining insights into
  training data and monitoring distribution shift.
\newblock In \emph{First International Workshop on Artificial Intelligence
  Safety Engineering}, 2018.

\bibitem[Biggio et~al.(2012)Biggio, Nelson, and
  Laskov]{10.5555/3042573.3042761}
Battista Biggio, Blaine Nelson, and Pavel Laskov.
\newblock Poisoning attacks against support vector machines.
\newblock In \emph{Proceedings of the 29th International Coference on
  International Conference on Machine Learning}, ICML’12, page 1467–1474,
  Madison, WI, USA, 2012. Omnipress.
\newblock ISBN 9781450312851.

\bibitem[Bishop et~al.(2014)Bishop, Bloomfield, Littlewood, Popov, Povyakalo,
  and Strigini]{bishop_conservative_2014}
Peter Bishop, Robin Bloomfield, Bev Littlewood, Peter Popov, Andrey Povyakalo,
  and Lorenzo Strigini.
\newblock A conservative bound for the probability of failure of a 1-out-of-2
  protection system with one hardware-only and one software-based protection
  train.
\newblock \emph{Reliability Engineering \& System Safety}, 130:\penalty0
  61--68, 2014.

\bibitem[Carlini et~al.(2019)Carlini, Liu, Erlingsson, Kos, and Song]{236216}
Nicholas Carlini, Chang Liu, {\'U}lfar Erlingsson, Jernej Kos, and Dawn Song.
\newblock The secret sharer: Evaluating and testing unintended memorization in
  neural networks.
\newblock In \emph{28th {USENIX} Security Symposium ({USENIX} Security 19)},
  pages 267--284, Santa Clara, CA, August 2019. {USENIX} Association.
\newblock ISBN 978-1-939133-06-9.
\newblock URL
  \url{https://www.usenix.org/conference/usenixsecurity19/presentation/carlini}.

\bibitem[Cheng(2020)]{cheng2020provablyrobust}
Chih-Hong Cheng.
\newblock Provably-robust runtime monitoring of neuron activation patterns,
  2020.

\bibitem[Cheng et~al.(2018)Cheng, N{\"{u}}hrenberg, and
  Yasuoka]{DBLP:journals/corr/abs-1809-06573}
Chih{-}Hong Cheng, Georg N{\"{u}}hrenberg, and Hirotoshi Yasuoka.
\newblock Runtime monitoring neuron activation patterns.
\newblock \emph{CoRR}, abs/1809.06573, 2018.
\newblock URL \url{http://arxiv.org/abs/1809.06573}.

\bibitem[Feldman(2019)]{DBLP:journals/corr/abs-1906-05271}
Vitaly Feldman.
\newblock Does learning require memorization? {A} short tale about a long tail.
\newblock \emph{CoRR}, abs/1906.05271, 2019.
\newblock URL \url{http://arxiv.org/abs/1906.05271}.

\bibitem[Fredrikson et~al.(2015)Fredrikson, Jha, and Ristenpart]{FJR2015}
Matt Fredrikson, Somesh Jha, and Thomas Ristenpart.
\newblock Model inversion attacks that exploit confidence information and basic
  countermeasures.
\newblock In \emph{CCS2015}, 2015.

\bibitem[Gehr et~al.(2018)Gehr, Mirman, Drachsler-Cohen, Tsankov, Chaudhuri,
  and Vechev]{gehr2018ai}
Timon Gehr, Matthew Mirman, Dana Drachsler-Cohen, Petar Tsankov, Swarat
  Chaudhuri, and Martin Vechev.
\newblock {AI2}: Safety and robustness certification of neural networks with
  abstract interpretation.
\newblock In \emph{Security and Privacy (SP), 2018 IEEE Symposium on}, 2018.

\bibitem[Gopinath et~al.(2018)Gopinath, Wang, Zhang, Pasareanu, and
  Khurshid]{DBLP:journals/corr/abs-1807-10439}
Divya Gopinath, Kaiyuan Wang, Mengshi Zhang, Corina~S. Pasareanu, and Sarfraz
  Khurshid.
\newblock Symbolic execution for deep neural networks.
\newblock \emph{CoRR}, abs/1807.10439, 2018.
\newblock URL \url{http://arxiv.org/abs/1807.10439}.

\bibitem[Guo et~al.(2018)Guo, Jiang, Zhao, Chen, and
  Sun]{10.1145/3236024.3264835}
Jianmin Guo, Yu~Jiang, Yue Zhao, Quan Chen, and Jiaguang Sun.
\newblock Dlfuzz: Differential fuzzing testing of deep learning systems.
\newblock In \emph{Proceedings of the 2018 26th ACM Joint Meeting on European
  Software Engineering Conference and Symposium on the Foundations of Software
  Engineering}, ESEC/FSE 2018, page 739–743, New York, NY, USA, 2018.
  Association for Computing Machinery.
\newblock ISBN 9781450355735.
\newblock \doi{10.1145/3236024.3264835}.
\newblock URL \url{https://doi.org/10.1145/3236024.3264835}.

\bibitem[Henzinger et~al.(2019)Henzinger, Lukina, and
  Schilling]{DBLP:journals/corr/abs-1911-09032}
Thomas~A. Henzinger, Anna Lukina, and Christian Schilling.
\newblock Outside the box: Abstraction-based monitoring of neural networks.
\newblock \emph{CoRR}, abs/1911.09032, 2019.
\newblock URL \url{http://arxiv.org/abs/1911.09032}.

\bibitem[{Huang} et~al.(2020){Huang}, {Zhou}, {Sun}, {Sharp}, {Maskell}, and
  {Huang}]{9340720}
W.~{Huang}, Y.~{Zhou}, Y.~{Sun}, J.~{Sharp}, S.~{Maskell}, and X.~{Huang}.
\newblock Practical verification of neural network enabled state estimation
  system for robotics.
\newblock In \emph{2020 IEEE/RSJ International Conference on Intelligent Robots
  and Systems (IROS)}, pages 7336--7343, 2020.
\newblock \doi{10.1109/IROS45743.2020.9340720}.

\bibitem[Huang et~al.(2020{\natexlab{a}})Huang, Sun, Sharp, Ruan, Meng, and
  Huang]{huang2020coverage}
Wei Huang, Youcheng Sun, James Sharp, Wenjie Ruan, Jie Meng, and Xiaowei Huang.
\newblock Coverage guided testing for recurrent neural networks,
  2020{\natexlab{a}}.

\bibitem[Huang et~al.(2017)Huang, Kwiatkowska, Wang, and Wu]{HKWW2017}
Xiaowei Huang, Marta Kwiatkowska, Sen Wang, and Min Wu.
\newblock Safety verification of deep neural networks.
\newblock In \emph{CAV2017}, pages 3--29, 2017.
\newblock \doi{10.1007/978-3-319-63387-9_1}.
\newblock URL \url{https://doi.org/10.1007/978-3-319-63387-9_1}.

\bibitem[Huang et~al.(2020{\natexlab{b}})Huang, Kroening, Ruan, Sharp, Sun,
  Thamo, Wu, and Yi]{HUANG2020100270}
Xiaowei Huang, Daniel Kroening, Wenjie Ruan, James Sharp, Youcheng Sun, Emese
  Thamo, Min Wu, and Xinping Yi.
\newblock A survey of safety and trustworthiness of deep neural networks:
  Verification, testing, adversarial attack and defence, and interpretability.
\newblock \emph{Computer Science Review}, 37:\penalty0 100270,
  2020{\natexlab{b}}.
\newblock ISSN 1574-0137.
\newblock \doi{https://doi.org/10.1016/j.cosrev.2020.100270}.
\newblock URL
  \url{http://www.sciencedirect.com/science/article/pii/S1574013719302527}.

\bibitem[Hyvärinen and Oja(2000)]{HYVARINEN2000411}
A.~Hyvärinen and E.~Oja.
\newblock Independent component analysis: algorithms and applications.
\newblock \emph{Neural Networks}, 13\penalty0 (4):\penalty0 411 -- 430, 2000.
\newblock ISSN 0893-6080.
\newblock \doi{https://doi.org/10.1016/S0893-6080(00)00026-5}.
\newblock URL
  \url{http://www.sciencedirect.com/science/article/pii/S0893608000000265}.

\bibitem[Jin et~al.(2020)Jin, Yi, Zhang, Zhang, Schewe, and Huang]{jin2020does}
Gaojie Jin, Xinping Yi, Liang Zhang, Lijun Zhang, Sven Schewe, and Xiaowei
  Huang.
\newblock How does weight correlation affect the generalisation ability of deep
  neural networks.
\newblock In \emph{NeurIPS'20}, 2020.

\bibitem[Katz et~al.(2017)Katz, Barrett, Dill, Julian, and
  Kochenderfer]{katz2017reluplex}
Guy Katz, Clark Barrett, David~L Dill, Kyle Julian, and Mykel~J Kochenderfer.
\newblock Reluplex: An efficient {SMT} solver for verifying deep neural
  networks.
\newblock In \emph{CAV2017}, pages 97--117, 2017.

\bibitem[Koller and Friedman(2009)]{PGM2009}
Daphne Koller and Nir Friedman.
\newblock \emph{Probabilistic Graphical Models}.
\newblock MIT Press, 2009.

\bibitem[Li et~al.(2019)Li, Liu, Yang, Chen, Huang, and
  Zhang]{10.1007/978-3-030-32304-2_15}
Jianlin Li, Jiangchao Liu, Pengfei Yang, Liqian Chen, Xiaowei Huang, and Lijun
  Zhang.
\newblock Analyzing deep neural networks with symbolic propagation: Towards
  higher precision and faster verification.
\newblock In Bor-Yuh~Evan Chang, editor, \emph{Static Analysis}, pages
  296--319, Cham, 2019. Springer International Publishing.
\newblock ISBN 978-3-030-32304-2.

\bibitem[Liu et~al.(2018)Liu, Ma, Aafer, Lee, Zhai, Wang, and
  Zhang]{LMALZWZ2018}
Yingqi Liu, Shiqing Ma, Yousra Aafer, Wen-Chuan Lee, Juan Zhai, Weihang Wang,
  and Xiangyu Zhang.
\newblock Trojaning attack on neural networks.
\newblock In \emph{Network and Distributed Systems Security (NDSS) Symposium
  2018}, 2018.

\bibitem[Lukina et~al.(2020)Lukina, Schilling, and
  Henzinger]{lukina2020unknown}
Anna Lukina, Christian Schilling, and Thomas~A. Henzinger.
\newblock Into the unknown: Active monitoring of neural networks, 2020.

\bibitem[Ma et~al.(2018)Ma, Juefei{-}Xu, Sun, Chen, Su, Zhang, Xue, Li, Li,
  Liu, Zhao, and Wang]{ma2018deepgauge}
Lei Ma, Felix Juefei{-}Xu, Jiyuan Sun, Chunyang Chen, Ting Su, Fuyuan Zhang,
  Minhui Xue, Bo~Li, Li~Li, Yang Liu, Jianjun Zhao, and Yadong Wang.
\newblock {DeepGauge}: Comprehensive and multi-granularity testing criteria for
  gauging the robustness of deep learning systems.
\newblock In \emph{ASE2018}, 2018.

\bibitem[Maaten and Hinton(2008)]{maaten2008visualizing}
Laurens van~der Maaten and Geoffrey Hinton.
\newblock {Visualizing data using t-SNE}.
\newblock \emph{Journal of machine learning research}, 9\penalty0
  (Nov):\penalty0 2579--2605, 2008.

\bibitem[Pei et~al.(2017)Pei, Cao, Yang, and Jana]{PCYJ2017}
Kexin Pei, Yinzhi Cao, Junfeng Yang, and Suman Jana.
\newblock {DeepXplore: Automated Whitebox Testing of Deep Learning Systems}.
\newblock In \emph{Proceedings of the 26th Symposium on Operating Systems
  Principles}, SOSP ’17, page 1–18, New York, NY, USA, 2017. Association
  for Computing Machinery.
\newblock ISBN 9781450350853.
\newblock \doi{10.1145/3132747.3132785}.
\newblock URL \url{https://doi.org/10.1145/3132747.3132785}.

\bibitem[Ruan et~al.(2018)Ruan, Huang, and Kwiatkowska]{RHK2018}
Wenjie Ruan, Xiaowei Huang, and Marta Kwiatkowska.
\newblock Reachability analysis of deep neural networks with provable
  guarantees.
\newblock In \emph{IJCAI}, pages 2651--2659, 2018.

\bibitem[Ruan et~al.(2019)Ruan, Wu, Sun, Huang, Kroening, and
  Kwiatkowska]{ijcai2019-824}
Wenjie Ruan, Min Wu, Youcheng Sun, Xiaowei Huang, Daniel Kroening, and Marta
  Kwiatkowska.
\newblock Global robustness evaluation of deep neural networks with provable
  guarantees for the hamming distance.
\newblock In \emph{Proceedings of the Twenty-Eighth International Joint
  Conference on Artificial Intelligence, {IJCAI-19}}, pages 5944--5952.
  International Joint Conferences on Artificial Intelligence Organization, 7
  2019.
\newblock \doi{10.24963/ijcai.2019/824}.
\newblock URL \url{https://doi.org/10.24963/ijcai.2019/824}.

\bibitem[Singh et~al.(2019)Singh, Gehr, P\"{u}schel, and
  Vechev]{Singh2019DeepPoly}
Gagandeep Singh, Timon Gehr, Markus P\"{u}schel, and Martin Vechev.
\newblock An abstract domain for certifying neural networks.
\newblock \emph{Proc. ACM Program. Lang.}, 3\penalty0 (POPL), January 2019.
\newblock \doi{10.1145/3290354}.
\newblock URL \url{https://doi.org/10.1145/3290354}.

\bibitem[{Sun} et~al.(2020){Sun}, {Zhou}, {Maskell}, {Sharp}, and
  {Huang}]{9196932}
Y.~{Sun}, Y.~{Zhou}, S.~{Maskell}, J.~{Sharp}, and X.~{Huang}.
\newblock Reliability validation of learning enabled vehicle tracking.
\newblock In \emph{2020 IEEE International Conference on Robotics and
  Automation (ICRA)}, pages 9390--9396, 2020.
\newblock \doi{10.1109/ICRA40945.2020.9196932}.

\bibitem[Sun et~al.(2018{\natexlab{a}})Sun, Huang, and
  Kroening]{DBLP:journals/corr/abs-1803-04792}
Youcheng Sun, Xiaowei Huang, and Daniel Kroening.
\newblock Testing deep neural networks.
\newblock \emph{CoRR}, abs/1803.04792, 2018{\natexlab{a}}.
\newblock URL \url{http://arxiv.org/abs/1803.04792}.

\bibitem[Sun et~al.(2018{\natexlab{b}})Sun, Wu, Ruan, Huang, Kwiatkowska, and
  Kroening]{sun2018concolic}
Youcheng Sun, Min Wu, Wenjie Ruan, Xiaowei Huang, Marta Kwiatkowska, and Daniel
  Kroening.
\newblock Concolic testing for deep neural networks.
\newblock In \emph{ASE}, 2018{\natexlab{b}}.

\bibitem[Sun et~al.(2019{\natexlab{a}})Sun, Huang, and
  Kroening]{sun2018testing}
Youcheng Sun, Xiaowei Huang, and Daniel Kroening.
\newblock Structural coverage metrics for deep neural networks.
\newblock \emph{EMSOFT2019}, 2019{\natexlab{a}}.

\bibitem[Sun et~al.(2019{\natexlab{b}})Sun, Huang, Kroening, Sharp, Hill, and
  Ashmore]{SHKSHA2019}
Youcheng Sun, Xiaowei Huang, Daniel Kroening, James Sharp, Matthew Hill, and
  Rob Ashmore.
\newblock Structural test coverage criteria for deep neural networks.
\newblock In \emph{Proceedings of the 41st International Conference on Software
  Engineering: Companion Proceedings}, ICSE '19, pages 320--321, Piscataway,
  NJ, USA, 2019{\natexlab{b}}. IEEE Press.
\newblock \doi{10.1109/ICSE-Companion.2019.00134}.
\newblock URL \url{https://doi.org/10.1109/ICSE-Companion.2019.00134}.

\bibitem[Szegedy et~al.(2014)Szegedy, Zaremba, Sutskever, Bruna, Erhan,
  Goodfellow, and Fergus]{szegedy2014intriguing}
Christian Szegedy, Wojciech Zaremba, Ilya Sutskever, Joan Bruna, Dumitru Erhan,
  Ian Goodfellow, and Rob Fergus.
\newblock Intriguing properties of neural networks.
\newblock In \emph{ICLR2014}, 2014.

\bibitem[Theodoridis and Koutroumbas(2003)]{theodoridis2003pattern}
Sergios Theodoridis and Konstantinos Koutroumbas.
\newblock \emph{Pattern recognition}.
\newblock Elsevier, 2003.

\bibitem[Weng et~al.(2018)Weng, Zhang, Chen, Yi, Su, Gao, Hsieh, and
  Daniel]{weng2018evaluating}
Tsui-Wei Weng, Huan Zhang, Pin-Yu Chen, Jinfeng Yi, Dong Su, Yupeng Gao,
  Cho-Jui Hsieh, and Luca Daniel.
\newblock Evaluating the robustness of neural networks: An extreme value theory
  approach, 2018.

\bibitem[Wicker et~al.(2018)Wicker, Huang, and
  Kwiatkowska]{10.1007/978-3-319-89960-2_22}
Matthew Wicker, Xiaowei Huang, and Marta Kwiatkowska.
\newblock Feature-guided black-box safety testing of deep neural networks.
\newblock In Dirk Beyer and Marieke Huisman, editors, \emph{Tools and
  Algorithms for the Construction and Analysis of Systems}, pages 408--426,
  Cham, 2018. Springer International Publishing.
\newblock ISBN 978-3-319-89960-2.

\bibitem[Wu et~al.(2018)Wu, Wicker, Ruan, Huang, and Kwiatkowska]{wu2018game}
Min Wu, Matthew Wicker, Wenjie Ruan, Xiaowei Huang, and Marta Kwiatkowska.
\newblock A game-based approximate verification of deep neural networks with
  provable guarantees.
\newblock \emph{arXiv preprint arXiv:1807.03571}, 2018.

\bibitem[Zhao et~al.(2020)Zhao, Banks, Sharp, Robu, Flynn, Fisher, and
  Huang]{zhao_safety_2020}
Xingyu Zhao, Alec Banks, James Sharp, Valentin Robu, David Flynn, Michael
  Fisher, and Xiaowei Huang.
\newblock A safety framework for critical systems utilising deep neural
  networks.
\newblock In \emph{SafeComp'20}, volume 12234 of \emph{{LNCS}}, pages 244--259,
  2020.

\bibitem[Zhao et~al.(2021)Zhao, Huang, Banks, Cox, Flynn, Schewe, and
  Huang]{zhao_safety_2021}
Xingyu Zhao, Wei Huang, Alec Banks, Victoria Cox, David Flynn, Sven Schewe, and
  Xiaowei Huang.
\newblock Assessing reliability of deep learning through robustness evaluation
  and operational testing.
\newblock In \emph{SafeComp'21}, volume 12234 of \emph{{LNCS}}, 2021.

\end{thebibliography}
